\documentclass[10pt,twocolumn,letterpaper]{article}
\usepackage[pagenumbers]{cvpr}

\usepackage{graphicx}
\usepackage{amsmath,amssymb}
\usepackage{booktabs}
\usepackage{multirow}
\usepackage{xcolor}
\usepackage{caption}
\usepackage{subcaption}
\usepackage{cuted}
\usepackage{microtype}

\definecolor{cvprblue}{rgb}{0.21,0.49,0.74}
\usepackage[pagebackref,breaklinks,colorlinks,allcolors=cvprblue]{hyperref}
\usepackage[capitalize]{cleveref}

\graphicspath{{figures/}}

\newcommand{\ageMax}{62}
\newcommand{\ageMean}{23.5}
\newcommand{\ageMedian}{20}
\newcommand{\ageMin}{18}
\newcommand{\ageSD}{8.0}
\newcommand{\aucCentre}{0.503}
\newcommand{\aucColour}{0.590}
\newcommand{\aucContrast}{0.598}
\newcommand{\aucEdges}{0.631}
\newcommand{\aucLum}{0.473}
\newcommand{\aucNear}{0.542}
\newcommand{\aucSI}{0.627}
\newcommand{\aucSOD}{0.534}
\newcommand{\aucSat}{0.539}

\newcommand{\bestEncoderAUC}{0.769}
\newcommand{\bestEncoderLabel}{DINOv3 ViT-B/16 patch-BERTScore F1}
\newcommand{\bestModel}{UNISAL}

\newcommand{\bestModelCC}{0.695}
\newcommand{\bestModelNSS}{1.27}

\newcommand{\cacClipAgree}{17}
\newcommand{\cacStyleP}{5\times10^{-9}}

\newcommand{\ccNine}{1.75}

\newcommand{\colSIpearson}{0.29}

\newcommand{\convImages}{260}
\newcommand{\distMean}{56.8}
\newcommand{\distPctRange}{90.4}
\newcommand{\distSD}{3.9}
\newcommand{\faceAreaPct}{5.1}
\newcommand{\faceBasePct}{7.2}
\newcommand{\faceEnrich}{5.0}
\newcommand{\faceLatency}{0.39}
\newcommand{\faceLatencyBase}{3.92}
\newcommand{\faceObsPct}{20.1}
\newcommand{\faceReachOne}{57}
\newcommand{\faceReachOneBase}{17}

\newcommand{\fdissAUC}{0.796}
\newcommand{\fdissPct}{99.6}

\newcommand{\fixDurMean}{433}
\newcommand{\fixDurMedian}{338}
\newcommand{\fixDurSD}{406}
\newcommand{\fixPerSp}{27.0}

\newcommand{\humanCC}{0.807}
\newcommand{\humanNSS}{1.58}

\newcommand{\imgColourMean}{36.3}
\newcommand{\imgColourSD}{17.2}
\newcommand{\imgSIMean}{59.8}
\newcommand{\imgSISD}{23.7}

\newcommand{\iocMean}{0.60}

\newcommand{\iocSD}{0.11}
\newcommand{\kNinetyFiveAUC}{3}
\newcommand{\kNinetyFiveCC}{6}

\newcommand{\kwStyleAgrP}{6\times10^{-9}}

\newcommand{\kwStyleEntP}{1\times10^{-7}}
\newcommand{\lineFollowP}{0.001}

\newcommand{\nAmateur}{5}
\newcommand{\nArtists}{81}
\newcommand{\nCountries}{13}
\newcommand{\nExpert}{1}
\newcommand{\nFaces}{287}
\newcommand{\nFemale}{42}
\newcommand{\nFixations}{121,426}
\newcommand{\nImages}{283}
\newcommand{\nIntraPairs}{12,735}
\newcommand{\nMale}{110}
\newcommand{\nMetaHigh}{58}
\newcommand{\nMetaLow}{67}
\newcommand{\nMetaMedium}{158}

\newcommand{\nPaintingsFaces}{93}
\newcommand{\nParticipants}{152}
\newcommand{\nScanpaths}{4,505}
\newcommand{\nssFirstSec}{3.57}
\newcommand{\nssLastSec}{1.50}
\newcommand{\paintingAreaPct}{65.6}

\newcommand{\pctCentralHalf}{62.8}
\newcommand{\pctFemale}{27.6}
\newcommand{\pctFrance}{85.5}
\newcommand{\pctHorizontal}{50.2}
\newcommand{\pctMale}{72.4}
\newcommand{\pctNovice}{96.1}

\newcommand{\pxPerDeg}{35.87}
\newcommand{\pxPerMM}{3.616}

\newcommand{\rhoAspectDisp}{0.78}
\newcommand{\rhoDiagCC}{0.16}
\newcommand{\rhoFacesDur}{0.38}
\newcommand{\rhoFacesEntropy}{-0.31}
\newcommand{\rhoHorizonEntropy}{0.34}
\newcommand{\saccAmpMean}{5.73}
\newcommand{\saccAmpMedian}{4.56}
\newcommand{\saccAmpSD}{4.32}
\newcommand{\screenWidthMM}{531}

\newcommand{\spPerImageMax}{32}
\newcommand{\spPerImageMedian}{15}
\newcommand{\spPerImageMin}{10}
\newcommand{\spiralEyeGain}{0.42}
\newcommand{\spiralEyeP}{0.009}

\newcommand{\uCompAgr}{0.036}

\newcommand{\uCompCen}{0.091}

\newcommand{\uCompEnt}{0.052}
\newcommand{\uContAgr}{0.023}

\newcommand{\uContDur}{0.107}

\newcommand{\uFmtAmp}{0.172}

\newcommand{\uFmtDisp}{0.301}

\newcommand{\uFmtEnt}{0.148}

\newcommand{\uStyAgr}{0.065}


\newcommand{\avatt}{AVAtt}
\newcommand{\dg}{\ensuremath{^\circ}}
\newcommand{\mypara}[1]{\vspace{2pt}\noindent\textbf{#1}\hspace{0.5em}}

\def\confName{CVPR}
\def\confYear{2026}

\title{\avatt: A Free-Viewing Eye-Tracking Dataset and Benchmark for Visual Attention on Paintings}

\author{Mohamed Amine Kerkouri\textsuperscript{1,}\thanks{Work done while at the University of Orl\'eans, PRISME
Laboratory. This paper extends our ETRA 2024 late-breaking work~\cite{kerkouri2024avatt} and the first author's PhD thesis~\cite{kerkouri2024thesis}.}\qquad Marouane Tliba\textsuperscript{2}\\
Aladine Chetouani\textsuperscript{2}\qquad Alessandro Bruno\textsuperscript{3}\\[4pt]
\textsuperscript{1}F-Initiatives, France\qquad
\textsuperscript{2}University of Orl\'eans, PRISME Laboratory, France\\
\textsuperscript{3}IULM University, Milan, Italy\\
{\tt\small m.a.kerkouri@f-initiatives.com}}

\begin{document}

\twocolumn[{%
\renewcommand\twocolumn[1][]{#1}%
\maketitle
\begin{center}
  \centering
  \includegraphics[width=0.97\textwidth]{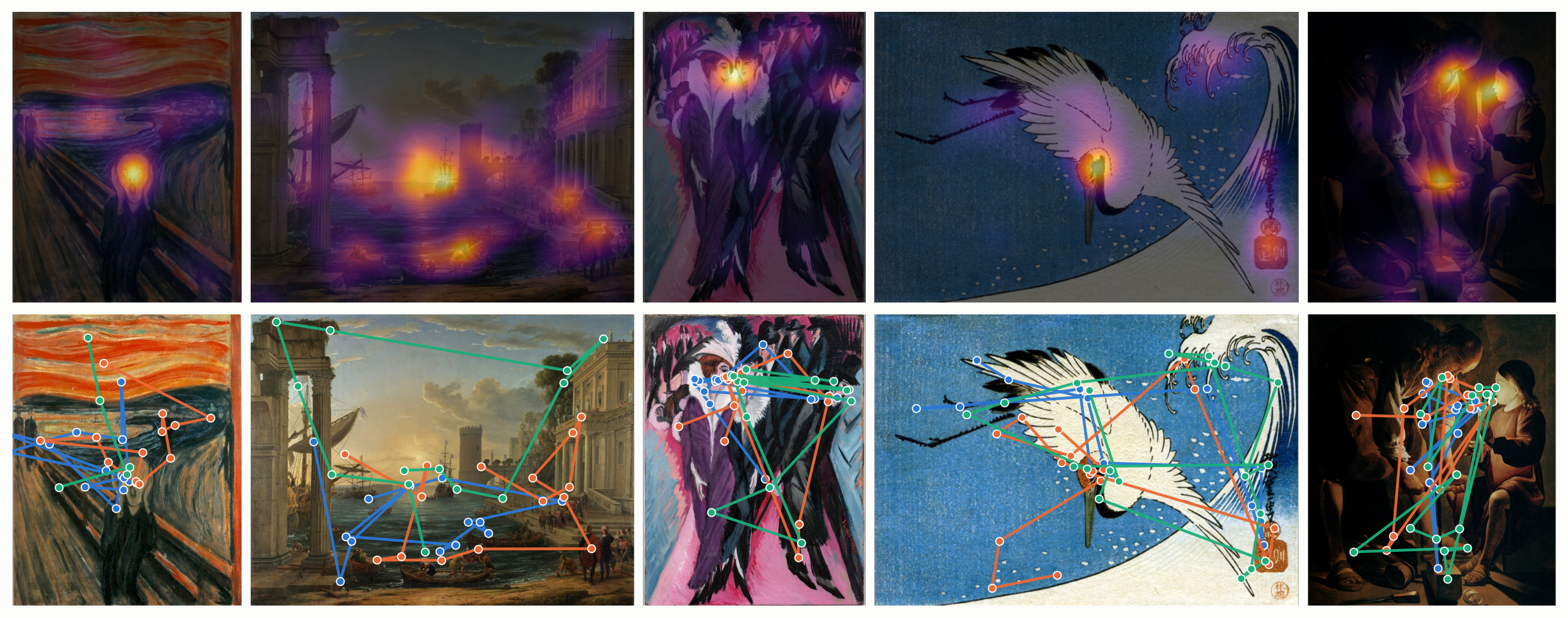}
  \captionof{figure}{\textbf{\avatt{} examples.} Top: saliency maps (one degree of visual angle) over the paintings.
  Bottom: three individual scanpaths (first 5\,s). \avatt{} provides individual scanpaths with timing and viewing
  distance, fixation and saliency maps, and temporal saliency videos for \nImages{} paintings from sixteen art
  movements.}
  \label{fig:teaser}
\end{center}
\vspace{6pt}
}]

\begin{abstract}
Saliency and scanpath models are developed almost exclusively on natural photographs, yet paintings are composed by
artists to guide the viewer's eye and differ from photographs in texture, colour and abstraction. We introduce
\avatt, a free-viewing eye-tracking dataset on \nImages{} paintings spanning sixteen art movements and five centuries,
with \nScanpaths{} scanpaths and \nFixations{} fixations from \nParticipants{} participants (\spPerImageMin--\spPerImageMax{}
observers per painting). For every painting we release individual scanpaths with fixation timing and eye--screen
distance, fixation maps, saliency maps smoothed at one degree of visual angle, temporal saliency videos and rich art
metadata. We benchmark classic and deep saliency models against human agreement, revealing substantial headroom for
art-aware models, and rank 44 scanpath-similarity measures, including a new language-free family of image-encoder
metrics built on ten CNN, ViT and self-supervised encoders. With controls that remove viewing biases, we show that
faces are fixated \faceEnrich$\times$ more than expected and within half a second, that local structure draws the eye,
that composition rules fitted from image content predict where people look, and that art style shapes how much viewers
agree. Dataset, code and a Python package are available at
\url{https://huggingface.co/datasets/Amine-CV/AVAtt}.
\end{abstract}

\section{Introduction}
\label{sec:intro}

Visual attention models are trained and evaluated on photographs of natural
scenes~\cite{judd2009mit,jiang2015salicon,borji2015cat2000}. Paintings challenge these models in several ways.
Their textures, palettes and degrees of abstraction differ from photographs, they span centuries of pictorial
conventions, and their content is \emph{composed}: artists place subjects, lines and contrasts deliberately to
organise how a picture is read. Eye tracking has been used to study the viewing of art since Yarbus'
recordings~\cite{yarbus1967,nodine1993art,quiroga2011art,massaro2012art,kerkouri2026divergent}, but public data remain scarce. The
largest public collection provides aggregated saliency maps for 150 paintings in five styles, without individual
scanpaths~\cite{lemeur2020paintings}, so scanpath models for paintings have had to rely on domain adaptation from photographs~\cite{kerkouri2022domain,tliba2022selfsupervised,kerkouri2026spgen}. Two factors explain the gap: recording enough observers per image is costly, and
copyright restricts the redistribution of artworks.

We present \avatt, a free-viewing eye-tracking dataset on paintings (\cref{fig:teaser}), first introduced as a
late-breaking result~\cite{kerkouri2024avatt}, together with benchmarks, analyses and an open toolkit. Our
contributions are:
\begin{itemize}
  \item \textbf{Dataset.} \nScanpaths{} scanpaths on \nImages{} paintings from sixteen movements, with fixation timing,
  per-fixation viewing distance, fixation and saliency maps derived from the measured viewing geometry, temporal
  saliency videos and art metadata (\cref{sec:dataset}).
  \item \textbf{Saliency benchmark.} Classic and deep models are evaluated on the painting area against human
  agreement, leaving clear headroom for art-aware models (\cref{sec:saliency}).
  \item \textbf{Scanpath-similarity benchmark.} 44 measures are ranked by how well they recognise viewers of the same
  painting. We introduce a language-free family of image-encoder metrics, and show that FDISS~\cite{kerkouri2026fdiss}
  and DINOv3-based matching lead (\cref{sec:scanpath}).
  \item \textbf{What draws attention in paintings.} Faces, local structure, composition and style are tested against
  a baseline that cancels central and scanning biases (\cref{sec:content}).
\end{itemize}
Data, code, the \texttt{avatt} Python package and all benchmark results are publicly
released.\footnote{Dataset: \url{https://huggingface.co/datasets/Amine-CV/AVAtt}; code:
\url{https://github.com/kmamine/AVAtt---Art-Visual-Attention}.}

\section{Related Work}
\label{sec:related}

\mypara{Eye-tracking datasets.} Free-viewing datasets such as MIT1003~\cite{judd2009mit},
CAT2000~\cite{borji2015cat2000} and the mouse-based SALICON~\cite{jiang2015salicon} drive saliency research
(\cref{tab:compare}). CAT2000 includes an art category among its twenty. \citet{lemeur2020paintings} collected
fixations on 150 paintings and showed that saliency models transfer imperfectly to art. Most other studies of viewing
art use small purpose-built stimulus sets and do not release data~\cite{nodine1993art,quiroga2011art,massaro2012art}.

\mypara{Saliency models.} Bottom-up models compute centre--surround contrast~\cite{itti1998model}, graph-based
activation~\cite{harel2006gbvs} or spectral residuals~\cite{hou2007spectral}. Deep models such as
UNISAL~\cite{droste2020unisal}, TranSalNet~\cite{lou2022transalnet}, SATSal~\cite{tliba2022satsal} and DeepGaze IIE~\cite{linardos2021deepgaze} lead
on natural images. Maps are scored with location- and distribution-based metrics~\cite{bylinskii2019metrics}.

\mypara{Scanpath prediction.} Scanpath models generate fixation sequences, from saccadic
models~\cite{lemeur2015saccadic} to deep networks such as DeepGaze III~\cite{kummerer2022deepgaze3},
SalyPath~\cite{kerkouri2021salypath,kerkouri2022salypath360} and lightweight or adversarially trained
predictors~\cite{kerkouri2022simple,kerkouri2023adversarial}. On paintings, they are adapted from photographs with
domain adaptation and self-supervision~\cite{kerkouri2022domain,tliba2022selfsupervised,kerkouri2026spgen}; \avatt{}
provides the per-observer art data needed to train and evaluate them directly.

\mypara{Scanpath similarity.} Scanpaths are compared with string-edit methods (ScanMatch~\cite{cristino2010scanmatch}),
geometric distances (DTW, Fr\'echet, Hausdorff, Mannan~\cite{mannan1995}), the five-dimensional
MultiMatch~\cite{dewhurst2012multimatch}, and FDISS~\cite{kerkouri2026fdiss}, which models fixations as foveal discs;
see~\cite{lemeur2013methods} for a review. Language-model-based metrics~\cite{kerkouri2026whatsaw,scanpathnlp} describe fixated regions with
a vision--language model and compare the descriptions with text metrics such as BERTScore~\cite{zhang2020bertscore}.

\mypara{Gaze as a probe.} Human gaze is also used to probe how vision models perceive, for masked
autoencoders~\cite{tliba2024mae}, quantised networks~\cite{kerkouri2024quantization} and foveated multimodal
LLMs~\cite{kerkouri2026foveated}, to guide segmentation~\cite{benyahia2026gazerefine}, and to characterise
observers~\cite{bruno2023cvd,wong2025shifts,kerkouri2026divergent}. Our image-encoder scanpath metrics follow this line
by reading gaze through the features of modern vision encoders.

\mypara{Composition.} Compositional devices such as the rule of thirds, golden sections and spirals, diagonals, and
Hambidge's dynamic symmetry~\cite{hambidge1926} are meant to organise attention. Photographic composition can be
classified automatically~\cite{lee2018kupcp,hong2021cacnet}, but the link between detected composition and gaze on
paintings has not been tested at scale.

\begin{table}[t]
  \centering
  \caption{Comparison with public free-viewing datasets.}
  \label{tab:compare}
  \resizebox{\columnwidth}{!}{%
  \begin{tabular}{llrcccc}
    \toprule
    Dataset & Domain & Images & Tracking & Scanpaths & Sal.\ maps & Temporal \\
    \midrule
    MIT1003~\cite{judd2009mit} & Natural & 1{,}003 & eye & \checkmark & \checkmark & -- \\
CAT2000~\cite{borji2015cat2000} & 20 categories & 4{,}000 & eye & \checkmark & \checkmark & -- \\
SALICON~\cite{jiang2015salicon} & Natural & 20{,}000 & mouse & \checkmark & \checkmark & -- \\
Le Meur \etal~\cite{lemeur2020paintings} & Paintings (5 styles) & 150 & eye & -- & \checkmark & -- \\
\textbf{AVAtt (ours)} & Paintings (16 movements) & 283 & eye & \checkmark & \checkmark & \checkmark \\ 
    \bottomrule
  \end{tabular}}
\end{table}

\section{The \avatt{} Dataset}
\label{sec:dataset}

\begin{figure}[t]
  \centering
  \includegraphics[width=\columnwidth]{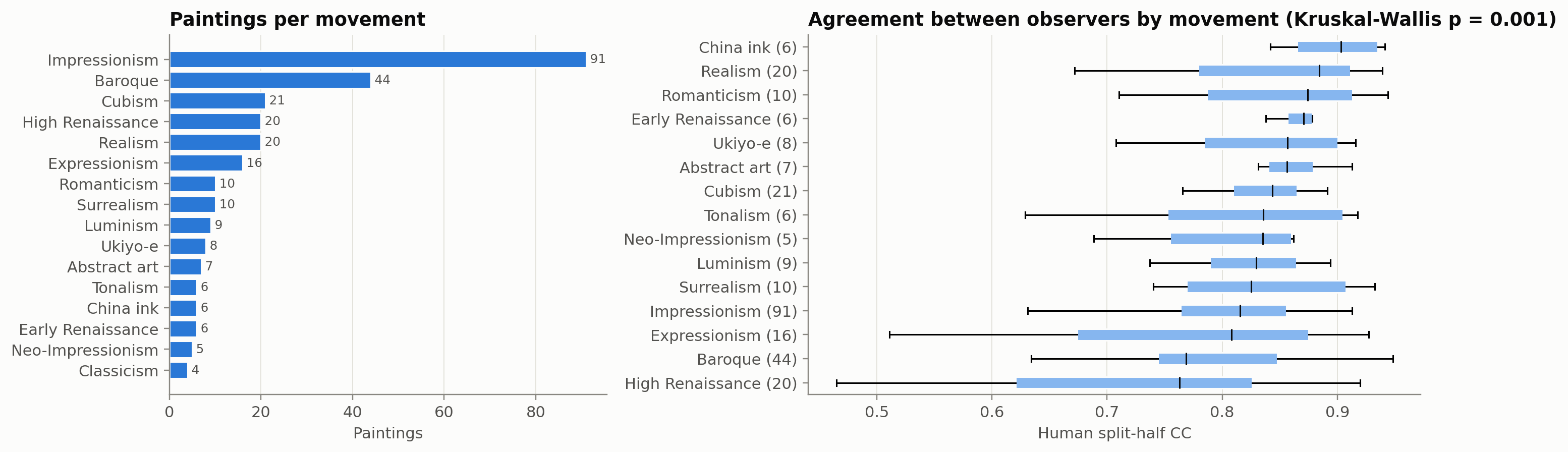}
  \caption{Paintings per movement group (left) and agreement between observers by movement (right).}
  \label{fig:movements}
\end{figure}

\mypara{Stimuli.} We collected 340 paintings from WikiArt~\cite{saleh2015wikiart}, Wikimedia Commons and other
open sources, covering sixteen movement groups: abstract art, Baroque, Chinese ink, Classicism, Cubism, Early
and High Renaissance, Expressionism, Impressionism, Luminism, Neo-Impressionism, Realism, Romanticism, Surrealism,
Tonalism and Ukiyo-e, from the 15th to the 21st century and by \nArtists{} identified artists
(\cref{fig:movements}). Each painting is annotated with artist, title, year, movement, genre and medium, with a
confidence level (\nMetaHigh{} works identified, \nMetaMedium{} attributed by artist, \nMetaLow{} annotated by movement
and genre only). Following~\cite{lemeur2020paintings}, each painting was resized without distortion and centred on a
1600$\times$900 grey canvas, which it covers by \paintingAreaPct\% on average. All image analyses in this paper are
restricted to the painting itself.

\mypara{Participants.} \nParticipants{} volunteers took part (\cref{tab:participants}): students, researchers and
staff, recruited by convenience sampling. Participants gave informed consent; the procedure followed the Declaration of
Helsinki~\cite{helsinki2013} and the EU GDPR, and the released data contain no personal information. All reported
normal or corrected-to-normal vision, and colour vision was screened with three Ishihara plates~\cite{ishihara1967}.

\begin{table}[t]
  \centering
  \caption{Participants (questionnaire data).}
  \label{tab:participants}
  \small
  \begin{tabular}{ll}
    \toprule
    Participants & \nParticipants \\
    Age (years) & \ageMean{} $\pm$ \ageSD{} (median \ageMedian, range \ageMin--\ageMax) \\
    Gender & \nMale{} men (\pctMale\%), \nFemale{} women (\pctFemale\%) \\
    Art expertise & \pctNovice\% novice, \nAmateur{} amateur, \nExpert{} expert \\
    Origin & \nCountries{} countries; France \pctFrance\% \\
    Vision & normal or corrected; Ishihara screening \\
    \bottomrule
  \end{tabular}
\end{table}

\mypara{Apparatus and protocol.} Gaze was recorded with a Tobii X2-30 eye tracker (30\,Hz) after 9-point calibration,
on a 1920$\times$1080 display in rooms with constant lighting. Participants were instructed to ``freely view and enjoy
the paintings''. Each painting was shown for 5.5\,s or 15\,s depending on the session, and was followed by a grey
screen of 1.5--2\,s. Fixations were extracted with the Tobii fixation filter (I-VT).

\mypara{Viewing geometry.} Rather than nominal values, we derive the geometry from the recordings. Regressing Tobii's
pixel gaze coordinates on its millimetre coordinates gives \pxPerMM\,px/mm (a \screenWidthMM\,mm-wide display). The
eye--screen distance recorded for each fixation averaged \distMean{} $\pm$ \distSD\,cm. One degree of visual angle
therefore spans $2d\tan(0.5\dg)\cdot\text{px/mm} = \pxPerDeg$\,px, which we use for all maps and regions.

\mypara{Processing and quality control.} Recordings were split by stimulus, and the grey inter-stimulus screens and
repeated viewings were discarded. Onsets are relative to the first fixation on the painting. A scanpath is kept if it
spans at least 4.5\,s, and a painting if at least ten valid scanpaths remain; ten observers suffice for stable maps
(\cref{sec:behaviour}). The released set has \nImages{} paintings, \nScanpaths{} scanpaths and \nFixations{} fixations,
with a median of \spPerImageMedian{} observers per painting.

\begin{figure}[t]
  \centering
  \includegraphics[width=\columnwidth]{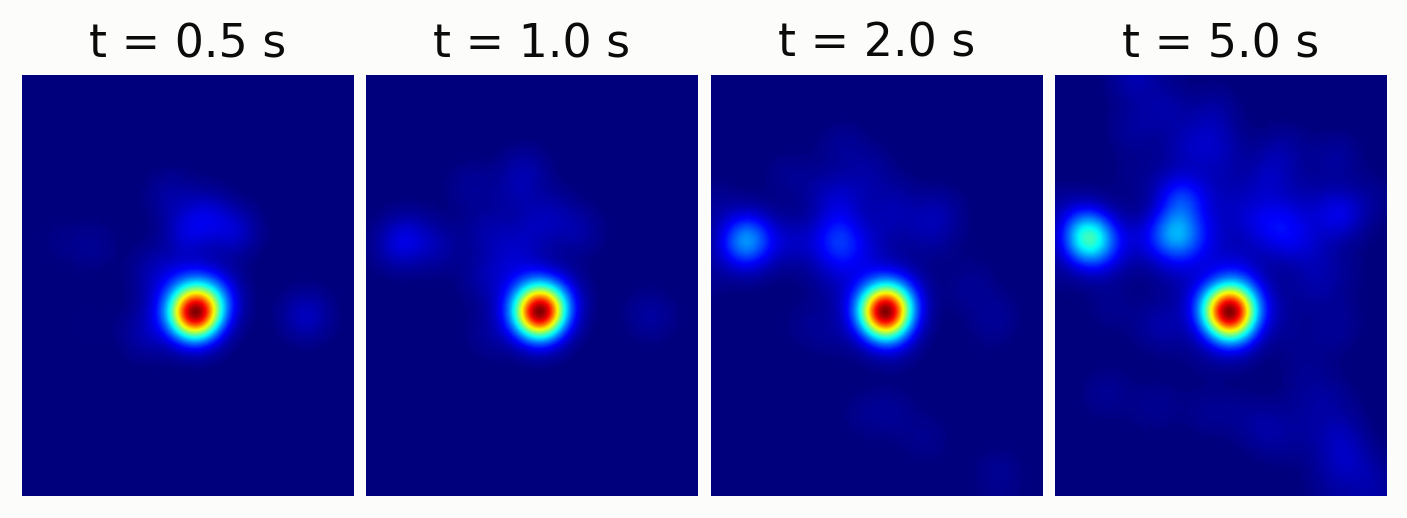}
  \caption{Frames of a temporal saliency video: attention accumulated up to $t$.}
  \label{fig:temporalframes}
\end{figure}

\mypara{Released data.} Each painting has a stable identifier (\texttt{avatt\_0001}--\texttt{avatt\_0283}) and
a metadata record (title, artist, year, movement, genre, medium, attribution confidence, copyright status, exposure,
observer count and painting box), together with:
(i) individual scanpaths as JSON lines, with per-fixation position, onset, duration and eye--screen distance;
(ii) a binary fixation map;
(iii) a saliency map, the fixation map convolved with a Gaussian of $\sigma=1\dg$; and
(iv) a temporal saliency video (60\,fps, 5\,s) in which every fixation contributes the time already spent on it,
$\min(\max(t-t_{\text{on}},0),\,d)$, so that each frame shows all attention accumulated up to time~$t$
(\cref{fig:temporalframes}).
The dataset is hosted on Hugging Face and loads with \texttt{datasets} in one line. The \texttt{avatt} package adds
the videos, map construction and the saliency metrics used in \cref{sec:saliency}, and the code repository reproduces
every table and figure. Gaze data and annotations are released under CC0, code under MIT.

\section{Viewing Behaviour on Paintings}
\label{sec:behaviour}

\begin{figure*}[t]
  \centering
  \includegraphics[width=\textwidth]{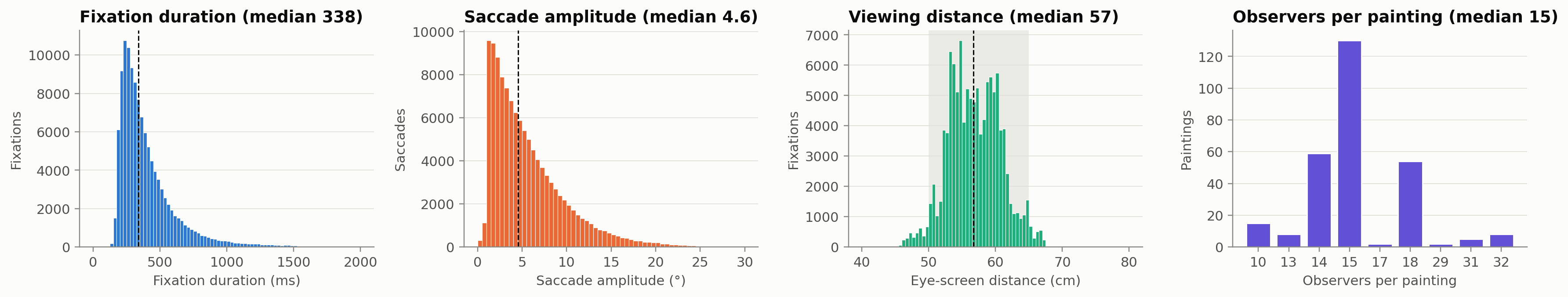}
  \caption{Fixation duration, saccade amplitude, eye--screen distance per fixation (shaded: 50--65\,cm) and observers
  per painting.}
  \label{fig:stats}
\end{figure*}

\mypara{Eye movements.} Scanpaths contain \fixPerSp{} fixations on average (\cref{fig:stats}). Fixation durations have
a median of \fixDurMedian\,ms (mean \fixDurMean{} $\pm$ \fixDurSD). Saccade amplitudes, computed with each fixation's
viewing distance, have a median of \saccAmpMedian\dg{} (mean \saccAmpMean{} $\pm$ \saccAmpSD). \pctHorizontal\% of
saccades lie within $\pm$30\dg{} of horizontal, against 33\% for uniform directions, and \pctCentralHalf\% of fixations
fall in the central half of the canvas, the usual central tendency~\cite{tatler2007central}.

\begin{figure*}[t]
  \centering
  \begin{subfigure}{0.49\textwidth}\includegraphics[width=\linewidth]{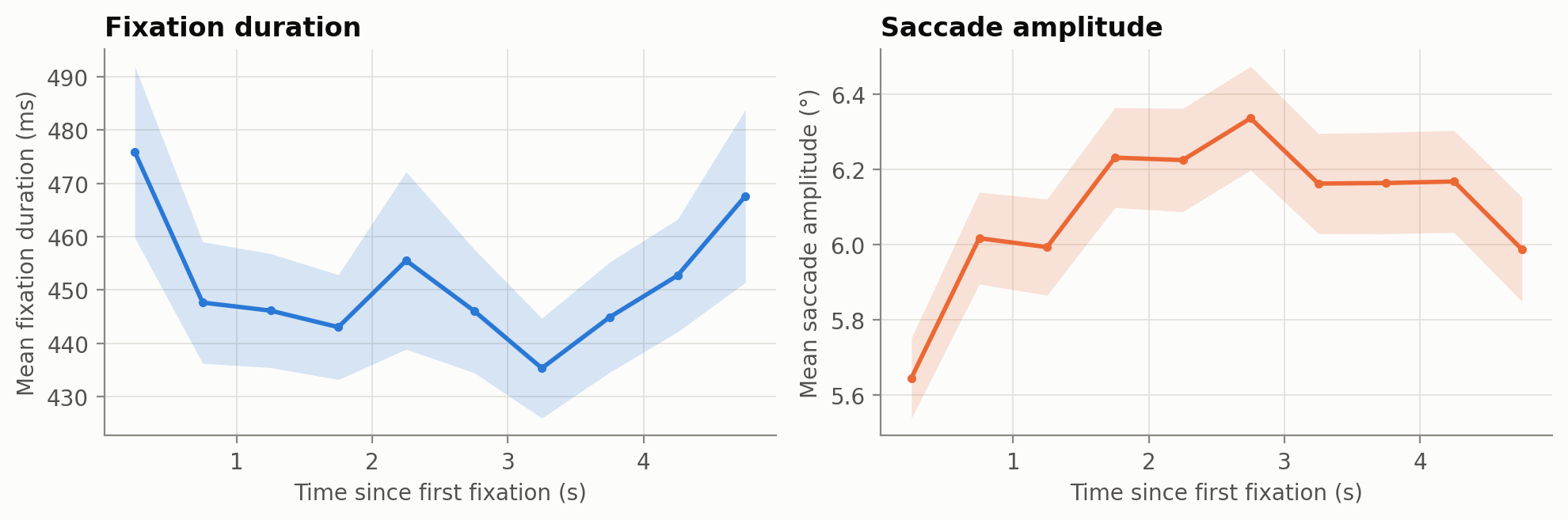}\end{subfigure}\hfill
  \begin{subfigure}{0.245\textwidth}\includegraphics[width=\linewidth]{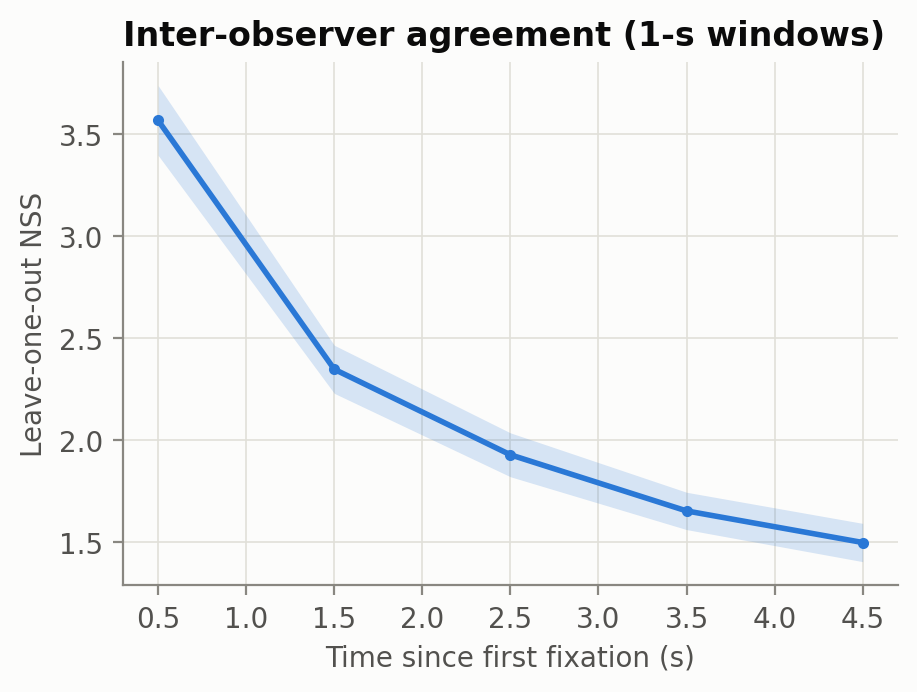}\end{subfigure}\hfill
  \begin{subfigure}{0.245\textwidth}\includegraphics[width=\linewidth]{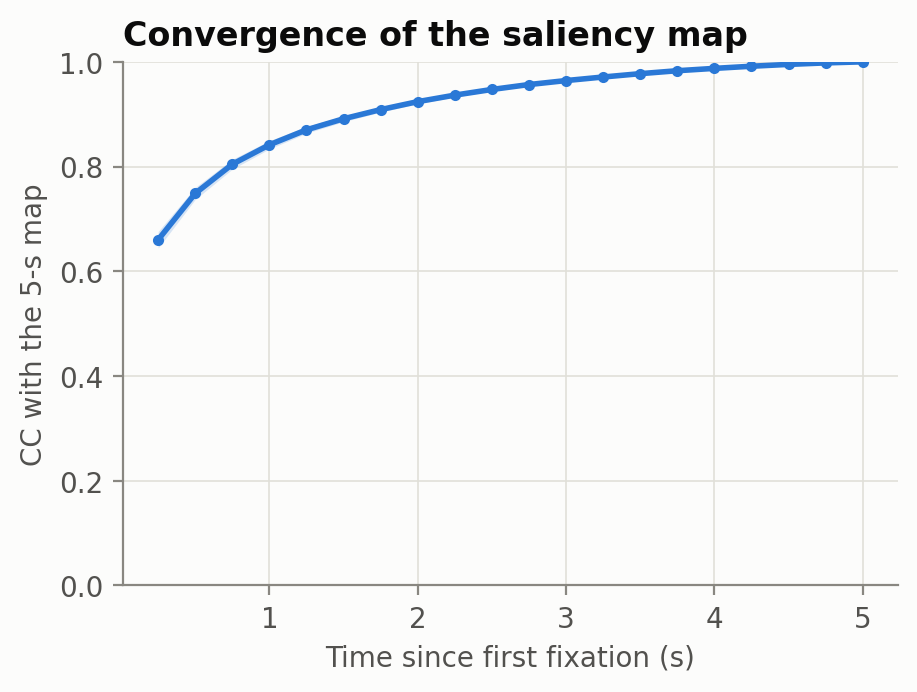}\end{subfigure}
  \caption{Temporal dynamics over the first 5\,s (mean $\pm$ 95\% CI): fixation duration and saccade amplitude,
  inter-observer agreement, and convergence of the cumulative saliency map.}
  \label{fig:temporal}
\end{figure*}

\mypara{Temporal dynamics.} Over the first five seconds, which every painting shares, viewing follows the
ambient-to-focal progression~\cite{pannasch2008ambient}: fixations lengthen and saccades shorten
(\cref{fig:temporal}). Agreement between observers, measured as the leave-one-out NSS in 1-s windows, falls from
\nssFirstSec{} in the first second to \nssLastSec{} in the fifth. The cumulative saliency map reaches a correlation of
0.9 with the 5-s map after \ccNine\,s.

\mypara{Inter-observer agreement.} Inter-observer congruency, the share of an observer's fixations
inside the Otsu-thresholded~\cite{otsu1979} map of the others, is \iocMean{} $\pm$ \iocSD. On the \convImages{} paintings with at least
fourteen observers, maps built from $k$ observers approach a held-out group of five quickly: AUC reaches 95\% of its
nine-observer value with \kNinetyFiveAUC{} observers, and CC and NSS with \kNinetyFiveCC{} (supp.\ \cref{fig:supp-convergence}).

\section{Benchmarks}
\label{sec:benchmarks}

\subsection{Saliency prediction}
\label{sec:saliency}

\mypara{Protocol.} Every predictor receives the painting crop and is evaluated on the painting area, using only
fixations on the painting. For each painting, observers are split ten times into random halves, and every predictor is
scored against the same held-out half with AUC-Judd, shuffled AUC (negatives: fixations from other paintings in
painting coordinates), NSS, CC, SIM and KL~\cite{bylinskii2019metrics}. We evaluate the map of the other half
(\emph{human} ceiling), a single observer, three deep models with their released weights
(DeepGaze IIE~\cite{linardos2021deepgaze} with its MIT1003 centre bias, TranSalNet-Res~\cite{lou2022transalnet},
UNISAL~\cite{droste2020unisal}), three classic models (GBVS~\cite{harel2006gbvs} and Itti--Koch~\cite{itti1998model}
from a PyTorch port of the GBVS toolbox~\cite{gbvstorch}, and the spectral residual~\cite{hou2007spectral}), the
salient-object map of BiRefNet~\cite{zheng2024birefnet}, and two centre biases: the fixations of all other paintings
in painting coordinates, and a Gaussian.

\mypara{Results.} Deep models clearly outperform the classic ones (\cref{tab:benchmark}), and \bestModel{} leads with
NSS \bestModelNSS{} and CC \bestModelCC. Human agreement (CC \humanCC, NSS \humanNSS) still sets a markedly higher
ceiling: paintings leave substantial headroom for art-aware saliency models, which \avatt{} is designed to train and
evaluate.

\begin{table}[t]
  \centering
  \caption{Saliency benchmark on the paintings (mean over paintings, split-half). Bold: best non-human predictor.}
  \label{tab:benchmark}
  \resizebox{\columnwidth}{!}{%
  \begin{tabular}{lcccccc}
    \toprule
    Predictor & AUC-J & sAUC & NSS & CC & SIM & KL$\downarrow$ \\
    \midrule
    Human (other half) & 0.807 & 0.713 & 1.577 & 0.807 & 0.722 & 0.551 \\
Single observer & 0.737 & 0.654 & 1.084 & 0.548 & 0.498 & 4.206 \\
\midrule
DeepGaze IIE & 0.771 & 0.620 & 1.217 & \textbf{0.695} & 0.638 & 0.589 \\
TranSalNet-Res & 0.771 & 0.649 & 1.205 & 0.680 & 0.658 & 0.533 \\
UNISAL & \textbf{0.776} & \textbf{0.654} & \textbf{1.270} & 0.695 & \textbf{0.659} & 0.600 \\
GBVS & 0.744 & 0.594 & 0.943 & 0.559 & 0.618 & \textbf{0.524} \\
Itti-Koch & 0.724 & 0.621 & 0.842 & 0.489 & 0.592 & 0.561 \\
Spectral residual & 0.670 & 0.629 & 0.632 & 0.355 & 0.542 & 0.717 \\
Salient objects (BiRefNet) & 0.573 & 0.534 & 0.393 & 0.204 & 0.281 & 14.073 \\
\midrule
Centre bias (dataset) & 0.718 & 0.503 & 0.774 & 0.490 & 0.590 & 0.572 \\
Centre bias (Gaussian) & 0.701 & 0.503 & 0.719 & 0.463 & 0.577 & 0.619 \\ 
    \bottomrule
  \end{tabular}}
\end{table}

\subsection{Scanpath similarity}
\label{sec:scanpath}

\mypara{Test design.} A good similarity measure should rate two observers of the same painting as more similar than
unrelated scanpaths. For each painting we sample up to 45 \emph{intra-image} pairs (\nIntraPairs{} in total). For each
pair we build an \emph{inter-image} baseline: the first observer against the scanpath of an observer of another
painting shown for the same duration, mapped onto the current painting by its relative position in the painting. The
baseline has realistic scanpath statistics and lies on the same painting, so content-based measures cannot exploit
differences between paintings. Scanpaths are cut to 5\,s. We report the pair-level AUC of intra versus inter pairs and
a paired Wilcoxon test over paintings, with the rank-biserial effect size and Benjamini--Hochberg
correction~\cite{benjamini1995fdr}.

\mypara{Spatial measures.} We test FDISS~\cite{kerkouri2026fdiss} (1\dg{} discs) and the eyefeatures~\cite{eyefeatures}
implementations of MultiMatch, DTW, discrete Fr\'echet, Hausdorff, Euclidean, Mannan, EyeAnalysis, time-delay embedding
and ScanMatch.

\mypara{Image-encoder measures.} Language-model-based metrics~\cite{kerkouri2026whatsaw,scanpathnlp} describe each fixated 96-px patch with
a vision--language model and compare paragraph descriptions with text metrics, at the cost of thousands of generative
calls per painting. We keep the idea of comparing \emph{what} was fixated, but replace the text with image embeddings.
Each painting is encoded once. A fixation is the mean of the dense feature map over its 96-px window ($\approx$2.7\dg),
L2-normalised. Sequences are compared by \emph{patch-BERTScore} (greedy cosine matching, F1 of precision and recall, as
in BERTScore~\cite{zhang2020bertscore}), \emph{embedding DTW} (order-aware), and mean-pooled cosine. We compare
supervised CNNs (VGG-16~\cite{simonyan2015vgg}, ResNet-50~\cite{he2016resnet}, ConvNeXt-T~\cite{liu2022convnext}),
the original ViT-B/16~\cite{dosovitskiy2021vit}, CLIP~\cite{radford2021clip}, and self-supervised MAE~\cite{he2022mae},
DINOv2~\cite{oquab2024dinov2}, DINOv3~\cite{simeoni2025dinov3}, I-JEPA~\cite{assran2023ijepa} and
VICReg~\cite{bardes2022vicreg}.

\mypara{Results.} FDISS is the most discriminative measure (AUC \fdissAUC; intra pairs are more similar in
\fdissPct\% of paintings; \cref{tab:scanpath,tab:encoders}), followed by our encoder measures, led by
\bestEncoderLabel{} (\bestEncoderAUC). Without any language model, the best encoders outperform every classic distance,
including MultiMatch and ScanMatch. Greedy matching beats order-aware DTW and mean pooling for every encoder:
\emph{what} viewers look at carries more of the shared signal than \emph{the order} in which they look.
Self-supervised dense features (DINOv3, I-JEPA, VICReg, DINOv2) are among the strongest backbones, making them a
natural choice for content-aware scanpath evaluation.

\begin{table}[t]
  \centering
  \caption{Spatial scanpath measures, intra vs inter pairs. $r_{rb}$: rank-biserial effect size; \% img.: paintings where
  intra pairs are more similar; $q$: BH-corrected.}
  \label{tab:scanpath}
  \resizebox{\columnwidth}{!}{%
  \begin{tabular}{lcrrcccc}
    \toprule
    Measure & & Intra & Inter & AUC & $r_{rb}$ & \% img. & $q$ \\
    \midrule
    FDISS & $\uparrow$ & 0.1225 & 0.04939 & 0.796 & 1.00 & 99.6 & $<10^{-3}$ \\
Mannan & $\downarrow$ & 0.008298 & 0.01289 & 0.691 & 0.97 & 95.4 & $<10^{-3}$ \\
EyeAnalysis & $\downarrow$ & 0.03062 & 0.04701 & 0.687 & 0.97 & 94.7 & $<10^{-3}$ \\
MultiMatch position & $\uparrow$ & 0.8319 & 0.803 & 0.625 & 0.93 & 90.5 & $<10^{-3}$ \\
ScanMatch & $\downarrow$ & 100.4 & 103.6 & 0.593 & 0.73 & 83.4 & $<10^{-3}$ \\
Hausdorff & $\downarrow$ & 0.2949 & 0.328 & 0.589 & 0.83 & 80.9 & $<10^{-3}$ \\
DTW & $\downarrow$ & 0.78 & 0.9308 & 0.581 & 0.72 & 78.1 & $<10^{-3}$ \\
Time-delay embedding & $\downarrow$ & 0.1206 & 0.1341 & 0.570 & 0.66 & 75.3 & $<10^{-3}$ \\
Euclidean & $\downarrow$ & 1.287 & 1.453 & 0.562 & 0.63 & 74.6 & $<10^{-3}$ \\
Discrete Fr\'echet & $\downarrow$ & 0.2074 & 0.2235 & 0.541 & 0.44 & 66.8 & $<10^{-3}$ \\
MultiMatch direction & $\uparrow$ & 0.8279 & 0.8196 & 0.539 & 0.46 & 68.2 & $<10^{-3}$ \\
MultiMatch shape & $\uparrow$ & 0.9453 & 0.9432 & 0.536 & 0.48 & 67.5 & $<10^{-3}$ \\
MultiMatch length & $\uparrow$ & 0.9426 & 0.9403 & 0.528 & 0.42 & 65.4 & $<10^{-3}$ \\
MultiMatch duration & $\uparrow$ & 0.6468 & 0.6449 & 0.506 & 0.10 & 52.3 & 0.147 \\ 
    \bottomrule
  \end{tabular}}
\end{table}

\begin{table}[t]
  \centering
  \caption{Image-encoder scanpath measures: AUC of intra vs inter pairs (all $q<10^{-3}$). Bold: best per column.}
  \label{tab:encoders}
  \resizebox{\columnwidth}{!}{%
  \begin{tabular}{llccc}
    \toprule
    Encoder & Training & BERTScore & DTW & Mean-pool \\
    \midrule
    VGG-16 & CNN (supervised) & 0.724 & 0.637 & 0.710 \\
ResNet-50 & CNN (supervised) & 0.753 & 0.665 & 0.728 \\
ConvNeXt-T & CNN (supervised) & 0.656 & 0.542 & 0.620 \\
ViT-B/16 original (Dosovitskiy et al.) & ViT (supervised) & 0.692 & 0.613 & 0.678 \\
CLIP ViT-B/16 (OpenAI) & ViT (language-supervised) & 0.694 & 0.600 & 0.672 \\
MAE ViT-B/16 & Self-supervised & 0.683 & 0.614 & 0.669 \\
DINOv2 ViT-B/14 & Self-supervised & 0.748 & 0.666 & 0.703 \\
DINOv3 ViT-B/16 & Self-supervised & \textbf{0.769} & \textbf{0.686} & 0.707 \\
I-JEPA ViT-H/14 & Self-supervised & 0.752 & 0.679 & 0.699 \\
VICReg ResNet-50 & Self-supervised & 0.753 & 0.657 & \textbf{0.728} \\ 
    \bottomrule
  \end{tabular}}
\end{table}

\section{What Draws Attention in Paintings}
\label{sec:content}

\mypara{A bias-free baseline.} Fixations concentrate in the centre and follow habitual scanning patterns, so comparing
fixated with random locations confounds content with viewing habits. We instead compare observed fixations with an
\emph{other-paintings baseline}: whole scanpaths recorded on other paintings, mapped into the current painting by
relative position. Content attracts attention if it receives more fixations than this baseline predicts. All image
analyses run on the painting crop and never use gaze. We segment salient objects with BiRefNet~\cite{zheng2024birefnet};
detect people, animals and objects with YOLO26 instance segmentation and YOLOE-26 open-vocabulary
segmentation~\cite{ultralytics,wang2025yoloe}; detect faces from pose keypoints and YOLOE-26; and estimate sky with
ADE20K semantic segmentation and relative depth with YOLO26. This yields \nFaces{} faces in \nPaintingsFaces{}
paintings and up to three focal subjects per painting.

\begin{figure}[t]
  \centering
  \includegraphics[width=\columnwidth]{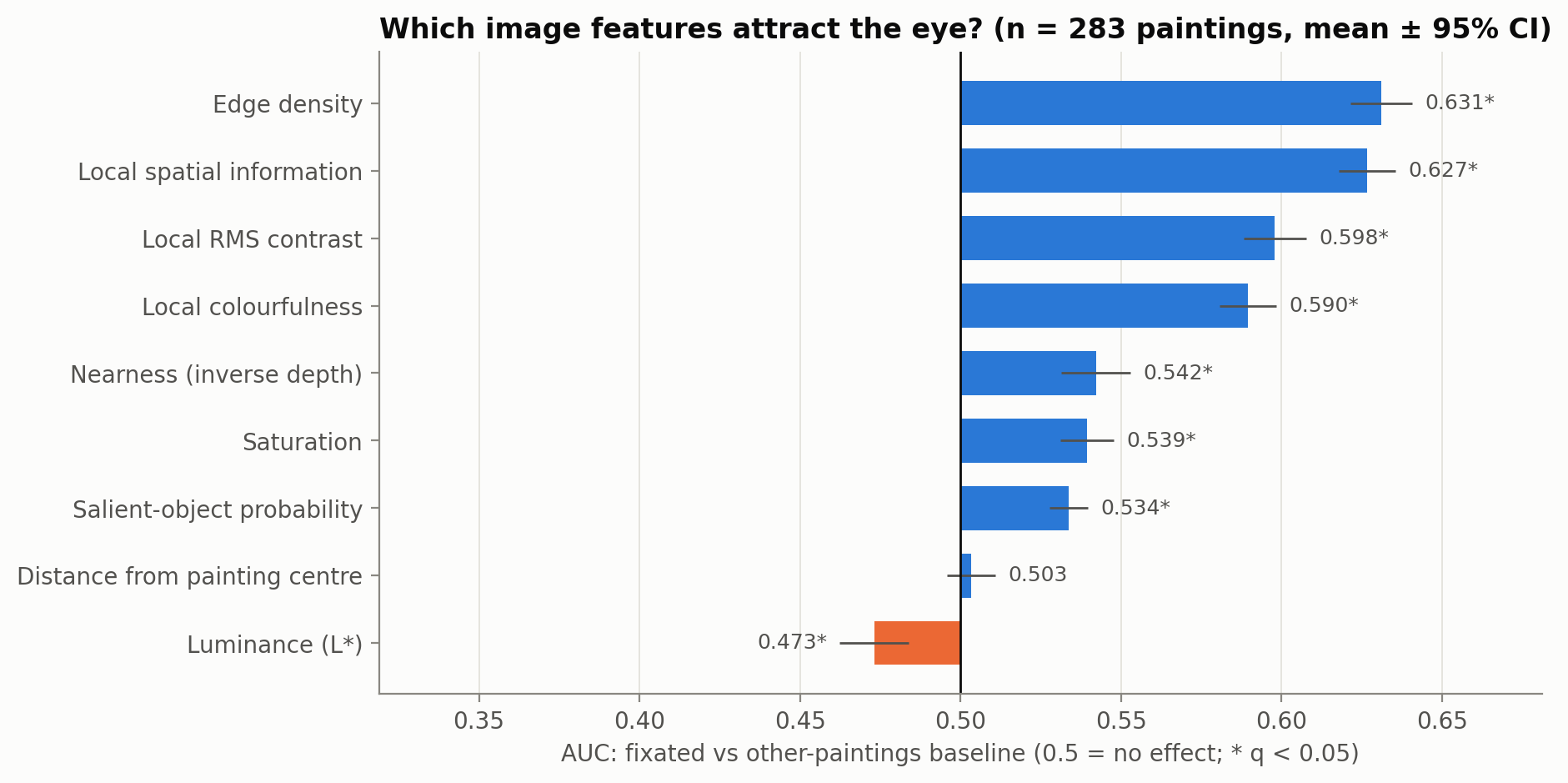}
  \caption{Image features at fixated vs baseline locations (AUC; * $q<0.05$).}
  \label{fig:features}
\end{figure}

\mypara{Image features.} We compare feature values at observed and baseline fixations (AUC; \cref{fig:features}).
Local structure attracts the eye most: edge density (\aucEdges), local spatial information (\aucSI), RMS contrast
(\aucContrast) and colourfulness (\aucColour). Nearness (\aucNear), saturation (\aucSat) and salient-object probability
(\aucSOD) add further positive effects, while bright regions are slightly avoided (\aucLum). Distance from the
painting centre is neutral (\aucCentre), confirming that the baseline fully absorbs the central tendency.

\mypara{Faces and objects.} Faces are the strongest attractor (\cref{tab:objects}). They cover \faceAreaPct\% of the
paintings that contain them but receive \faceObsPct\% of fixations, against \faceBasePct\% under the baseline. The median
first fixation on a face occurs at \faceLatency\,s (baseline \faceLatencyBase\,s), and \faceReachOne\% of viewers fixate
a face within one second (baseline \faceReachOneBase\%), as reported for photographs~\cite{cerf2008faces}. People,
vessels, animals and food are also over-fixated, whereas sky is under-fixated.

\begin{table}[t]
  \centering
  \caption{Attention to object categories. Enrich.: observed / baseline share of fixations; @1s: \% of viewers reaching
  the category within 1\,s, observed / baseline.}
  \label{tab:objects}
  \resizebox{\columnwidth}{!}{%
  \begin{tabular}{lrrrrrrr}
    \toprule
    Category & Paint. & Area\% & Fix.\% & Base\% & Enrich. & @1s & $q$ \\
    \midrule
    face & 93 & 5.1 & 20.1 & 7.2 & 4.97 & 57 / 17 & $<10^{-3}$ \\
person & 189 & 14.0 & 26.7 & 18.5 & 3.12 & 50 / 30 & $<10^{-3}$ \\
vessel / vehicle & 51 & 4.9 & 10.2 & 5.7 & 2.67 & 22 / 11 & $<10^{-3}$ \\
animal & 89 & 11.2 & 21.0 & 14.4 & 2.26 & 40 / 24 & $<10^{-3}$ \\
food / fruit & 13 & 3.2 & 7.2 & 5.1 & 1.94 & 22 / 16 & $<10^{-3}$ \\
building & 7 & 42.7 & 56.0 & 53.6 & 1.28 & 91 / 73 & 0.701 \\
tree / plant & 20 & 16.9 & 19.2 & 19.0 & 1.13 & 34 / 34 & 0.812 \\
water & 24 & 63.4 & 63.6 & 67.1 & 0.89 & 79 / 78 & 0.701 \\
sky & 32 & 52.3 & 26.9 & 50.6 & 0.47 & 52 / 74 & $<10^{-3}$ \\ 
    \bottomrule
  \end{tabular}}
\end{table}

\mypara{Composition.} We consider eight families of templates, defined in painting coordinates
(supp.\ \cref{fig:supp-templates}): the rule of thirds, the golden grid, the centre, the diagonals, golden triangles,
Hambidge's dynamic symmetry~\cite{hambidge1926}, and the golden spiral. The spiral has eight variants (four corners,
two orientations), either stretched to the canvas or as a true golden rectangle. Each variant is fitted to a painting
from image content only, combining the proximity of focal subjects to its key points with salient-object-weighted edge
strength along its lines, and scored against every other painting's content on the same template, which corrects for
the choice among variants (details in the supplementary material). Gaze is never used for fitting.

\mypara{Composition guides gaze.} The fit predicts where people look. Paintings that fit a family better receive more
fixations on its points and lines relative to the bias-free baseline, for \emph{every} family (Spearman
$\rho=0.12$--$0.29$; \cref{tab:composition}). The selected spiral variant attracts more fixations than the other seven
($p<10^{-3}$), the eye of the best true-ratio spiral is reached \spiralEyeGain\,s earlier than the baseline predicts
($p=\spiralEyeP$), and saccades align with the painting's dominant lines ($p=\lineFollowP$). Composition thus steers
attention by placing salient content on its points and lines. Because photographic composition
classifiers~\cite{hong2021cacnet,lee2018kupcp} and CLIP zero-shot agree on the top class for only \cacClipAgree\% of
paintings, \avatt{} also offers a gaze-grounded testbed for art-specific composition models.

\begin{table}[t]
  \centering
  \caption{Composition families. Var.: variants; Top / Bottom: fixation enrichment on the template (observed /
  baseline) for the best- and worst-fitting quarter of paintings; $\rho$: correlation between image-only fit and
  enrichment.}
  \label{tab:composition}
  \resizebox{\columnwidth}{!}{%
  \begin{tabular}{lrcccr}
    \toprule
    Family & Var. & Top & Bottom & $\rho$ & $p_\rho$ \\
    \midrule
    Rule of thirds & 1 & 1.05 & 0.89 & 0.22 & $<10^{-3}$ \\
Golden grid & 1 & 1.04 & 0.88 & 0.23 & $<10^{-3}$ \\
Centre & 1 & 0.77 & 0.63 & 0.12 & 0.075 \\
Diagonal & 2 & 1.18 & 0.93 & 0.29 & $<10^{-3}$ \\
Golden triangles & 2 & 1.13 & 0.99 & 0.19 & 0.001 \\
Dynamic symmetry & 1 & 1.02 & 0.97 & 0.12 & 0.044 \\
Golden spiral (stretched) & 8 & 1.21 & 0.84 & 0.22 & $<10^{-3}$ \\
Golden spiral (true ratio) & 8 & 1.21 & 0.94 & 0.26 & $<10^{-3}$ \\ 
    \bottomrule
  \end{tabular}}
\end{table}

\mypara{Painting-level factors.} Across paintings (supp.\ \cref{fig:supp-factors}), faces focus attention: more faces
mean lower saliency-map entropy ($\rho=\rhoFacesEntropy$) and longer fixations ($\rho=\rhoFacesDur$). Horizons spread
attention ($\rho=\rhoHorizonEntropy$), and diagonal lines increase agreement between observers ($\rho=\rhoDiagCC$).
Wide paintings produce more dispersed fixations ($\rho=\rhoAspectDisp$).

\subsection{Style, composition and content}
\label{sec:style}

\mypara{Style shapes viewing.} We compare the fifteen movement groups with at least five paintings, measuring viewing
over the first 5\,s only and adjusting for exposure duration. Agreement between observers, saliency-map entropy,
dispersion, fixation duration and saccade amplitude all differ across styles (Kruskal--Wallis $p=\kwStyleAgrP$ for
agreement, $p=\kwStyleEntP$ for entropy; supp.\ \cref{fig:supp-style-attention}). Chinese ink, Realism, Romanticism and Early Renaissance paintings produce the
most consistent viewing; High Renaissance, Impressionist and Expressionist paintings the least.

\mypara{Composition and content differ by style.} Styles favour different compositions
(supp.\ \cref{fig:supp-style-composition}): image-only fit scores differ across styles for the centre, golden-grid and
spiral templates (BH $q<0.05$), centred compositions dominate Cubist and Expressionist works, and the CACNet class
distribution depends on style ($\chi^2$ $p=\cacStyleP$). Content differs too (supp.\ \cref{fig:supp-style-objects}):
faces appear in every Early Renaissance painting but in few Impressionist ones, and animals and vessels concentrate in
Baroque works. Where faces are present they are over-fixated in every style (3--10$\times$): their pull is
universal across five centuries of painting. The link between composition fit and fixations holds within most styles,
so it does not merely reflect differences between styles.

\begin{figure}[t]
  \centering
  \includegraphics[width=\columnwidth]{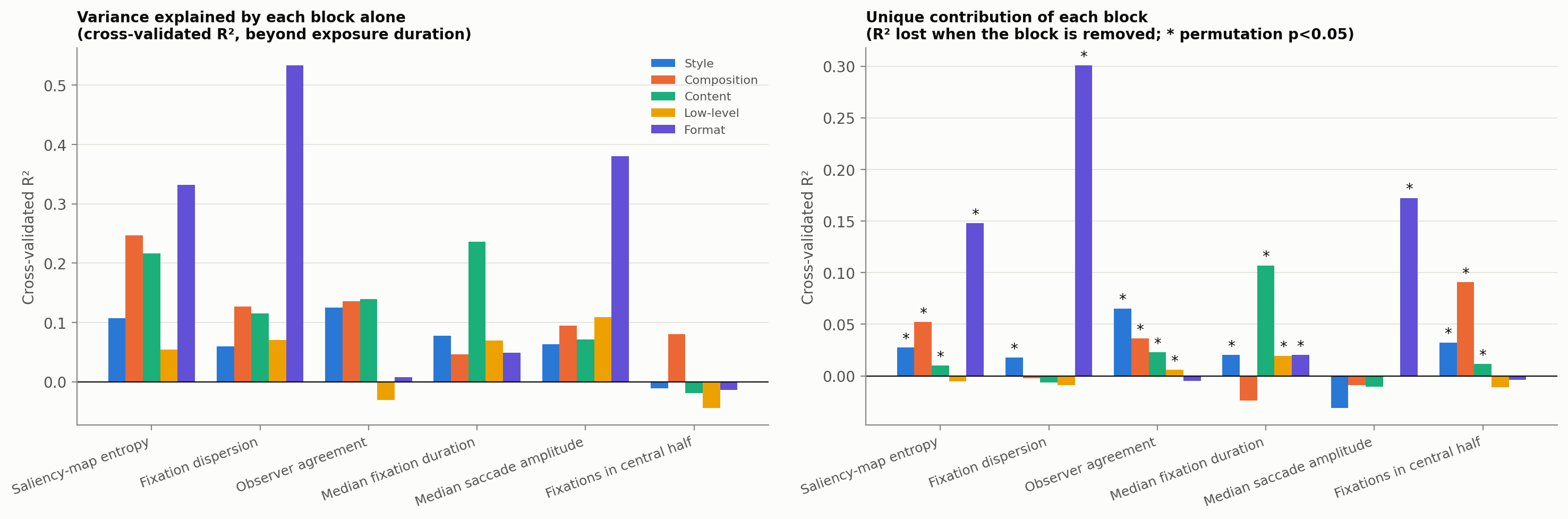}
  \caption{Variance in viewing explained by style, composition, content, low-level statistics and format
  (cross-validated $R^2$ beyond exposure duration). Left: each block alone. Right: unique contribution
  (* permutation $p<0.05$).}
  \label{fig:style-variance}
\end{figure}

\mypara{Which factor matters most?} To separate the factors, we predict each viewing measure from five blocks of
painting descriptors: \emph{style} (movement group), \emph{composition} (eight template fits and CACNet class
probabilities), \emph{content} (faces, focal subjects, salient-object area, horizon and object-category areas),
\emph{low-level} statistics (colourfulness, spatial information, symmetry, line orientations) and \emph{format} (aspect
ratio and on-screen size). We use ridge regression with cross-validated $R^2$ (10$\times$5-fold), always including
exposure duration. A block's \emph{unique} contribution is the $R^2$ lost when it is removed from the full model, and
is tested by permuting the block (200 permutations). The results (\cref{fig:style-variance}) depend on the measure:
\begin{itemize}
  \item \emph{Format} dominates the spatial spread of viewing: dispersion (unique $R^2$ \uFmtDisp), saccade amplitude
  (\uFmtAmp) and entropy (\uFmtEnt).
  \item \emph{Content} dominates fixation duration (\uContDur): faces and subjects hold the eye.
  \item \emph{Style} has the largest unique share of agreement between observers (\uStyAgr, ahead of composition
  \uCompAgr{} and content \uContAgr).
  \item \emph{Composition} is the only block that predicts the central tendency (\uCompCen) and adds to entropy
  (\uCompEnt) beyond everything else.
\end{itemize}
In short, \emph{where} the eye travels is set mostly
by the canvas format and the composition, \emph{how long} it dwells by content, and \emph{how much viewers agree} by
style.

\section{Conclusion}
\label{sec:conclusion}

\avatt{} provides individual scanpaths, saliency maps, temporal saliency videos and art metadata for \nImages{}
paintings from sixteen movements, together with benchmarks for saliency and scanpath models and an analysis of what
guides attention in art. Paintings leave clear headroom for saliency models; FDISS and our language-free encoder
metrics capture shared viewing best; and faces, local structure, composed salient content and style shape where and
how people look. Future work includes viewers with art expertise and diverse viewer populations~\cite{nodine1993art,kerkouri2026divergent}, higher-frequency recordings for
saccade dynamics, and art-aware saliency, scanpath and composition models trained on \avatt. Data, code and the
\texttt{avatt} package are available at \url{https://huggingface.co/datasets/Amine-CV/AVAtt} and
\url{https://github.com/kmamine/AVAtt---Art-Visual-Attention}.

\mypara{Acknowledgements.} This work was funded by the TIC-ART project of the Centre-Val de Loire Region, France.

{
  \small
  \bibliographystyle{ieeenat_fullname}
  \bibliography{refs}

@inproceedings{kerkouri2024avatt,
  title     = {{AVAtt}: Art Visual Attention dataset for diverse painting styles},
  author    = {Kerkouri, Mohamed Amine and Tliba, Marouane and Chetouani, Aladine and Bruno, Alessandro},
  booktitle = {Proceedings of the 2024 Symposium on Eye Tracking Research and Applications (ETRA '24)},
  pages     = {1--3},
  year      = {2024},
  publisher = {ACM},
  doi       = {10.1145/3649902.3655656}
}

@inproceedings{kerkouri2026fdiss,
  title     = {Closing the Foveal Gap: Perceptually Grounded Scanpath Comparison with Disc {IoU}},
  author    = {Kerkouri, Mohamed Amine and Tliba, Marouane and Sellam, Zakaria and Distante, Cosimo and Bruno, Alessandro and Chetouani, Aladine},
  booktitle = {Proceedings of the 2026 Symposium on Eye Tracking Research and Applications (ETRA '26)},
  pages     = {1--3},
  year      = {2026},
  publisher = {ACM},
  doi       = {10.1145/3797246.3805860}
}

@misc{eyefeatures,
  author = {Daudov, Vagiz},
  title  = {eyefeatures: feature engineering and scanpath distances for eye-tracking data},
  year   = {2026},
  note   = {Python package, version 2.0.2},
  howpublished = {\url{https://pypi.org/project/eyefeatures/}}
}

@misc{scanpathnlp,
  author = {Kerkouri, Mohamed Amine},
  title  = {scanpath-nlp-metrics: {NLP} and spatial metrics for scanpath comparison using vision-language models},
  year   = {2026},
  note   = {Python package, version 0.0.1},
  howpublished = {\url{https://pypi.org/project/scanpath-nlp-metrics/}}
}

@article{lemeur2020paintings,
  title   = {Can we accurately predict where we look at paintings?},
  author  = {Le Meur, Olivier and Le Pen, Tugdual and Cozot, R{\'e}mi},
  journal = {PLOS ONE},
  volume  = {15},
  number  = {10},
  pages   = {e0239980},
  year    = {2020},
  doi     = {10.1371/journal.pone.0239980}
}

@article{lemeur2015saccadic,
  title   = {Saccadic model of eye movements for free-viewing condition},
  author  = {Le Meur, Olivier and Liu, Zhi},
  journal = {Vision Research},
  volume  = {116},
  pages   = {152--164},
  year    = {2015}
}

@article{lemeur2013methods,
  title   = {Methods for comparing scanpaths and saliency maps: strengths and weaknesses},
  author  = {Le Meur, Olivier and Baccino, Thierry},
  journal = {Behavior Research Methods},
  volume  = {45},
  number  = {1},
  pages   = {251--266},
  year    = {2013}
}

@book{yarbus1967,
  title     = {Eye Movements and Vision},
  author    = {Yarbus, Alfred L.},
  publisher = {Plenum Press},
  address   = {New York},
  year      = {1967}
}

@article{massaro2012art,
  title   = {When art moves the eyes: a behavioral and eye-tracking study},
  author  = {Massaro, Davide and Savazzi, Federica and Di Dio, Cinzia and Freedberg, David and Gallese, Vittorio and Gilli, Gabriella and Marchetti, Antonella},
  journal = {PLOS ONE},
  volume  = {7},
  number  = {5},
  pages   = {e37285},
  year    = {2012}
}

@article{quiroga2011art,
  title   = {How do we see art: an eye-tracker study},
  author  = {Quiroga, Rodrigo Quian and Pedreira, Carlos},
  journal = {Frontiers in Human Neuroscience},
  volume  = {5},
  pages   = {98},
  year    = {2011}
}

@article{nodine1993art,
  title   = {The role of formal art training on perception and aesthetic judgment of art compositions},
  author  = {Nodine, Calvin F. and Locher, Paul J. and Krupinski, Elizabeth A.},
  journal = {Leonardo},
  volume  = {26},
  number  = {3},
  pages   = {219--227},
  year    = {1993}
}

@inproceedings{judd2009mit,
  title     = {Learning to predict where humans look},
  author    = {Judd, Tilke and Ehinger, Krista and Durand, Fr{\'e}do and Torralba, Antonio},
  booktitle = {IEEE International Conference on Computer Vision (ICCV)},
  pages     = {2106--2113},
  year      = {2009}
}

@inproceedings{jiang2015salicon,
  title     = {{SALICON}: Saliency in context},
  author    = {Jiang, Ming and Huang, Shengsheng and Duan, Juanyong and Zhao, Qi},
  booktitle = {IEEE Conference on Computer Vision and Pattern Recognition (CVPR)},
  pages     = {1072--1080},
  year      = {2015}
}

@article{borji2015cat2000,
  title   = {{CAT2000}: A large scale fixation dataset for boosting saliency research},
  author  = {Borji, Ali and Itti, Laurent},
  journal = {arXiv preprint arXiv:1505.03581},
  year    = {2015}
}

@article{bylinskii2019metrics,
  title   = {What do different evaluation metrics tell us about saliency models?},
  author  = {Bylinskii, Zoya and Judd, Tilke and Oliva, Aude and Torralba, Antonio and Durand, Fr{\'e}do},
  journal = {IEEE Transactions on Pattern Analysis and Machine Intelligence},
  volume  = {41},
  number  = {3},
  pages   = {740--757},
  year    = {2019}
}

@article{tatler2007central,
  title   = {The central fixation bias in scene viewing: Selecting an optimal viewing position independently of motor biases and image feature distributions},
  author  = {Tatler, Benjamin W.},
  journal = {Journal of Vision},
  volume  = {7},
  number  = {14},
  pages   = {4},
  year    = {2007}
}

@article{itti1998model,
  title   = {A model of saliency-based visual attention for rapid scene analysis},
  author  = {Itti, Laurent and Koch, Christof and Niebur, Ernst},
  journal = {IEEE Transactions on Pattern Analysis and Machine Intelligence},
  volume  = {20},
  number  = {11},
  pages   = {1254--1259},
  year    = {1998}
}

@inproceedings{harel2006gbvs,
  title     = {Graph-based visual saliency},
  author    = {Harel, Jonathan and Koch, Christof and Perona, Pietro},
  booktitle = {Advances in Neural Information Processing Systems (NIPS)},
  volume    = {19},
  year      = {2006}
}

@misc{gbvstorch,
  title        = {{GBVS\_Pytorch}: Graph-Based Visual Saliency in {PyTorch}},
  author       = {Kerkouri, Mohamed Amine},
  year         = {2026},
  howpublished = {\url{https://github.com/kmamine/GBVS_Pytorch}}
}

@inproceedings{hou2007spectral,
  title     = {Saliency detection: A spectral residual approach},
  author    = {Hou, Xiaodi and Zhang, Liqing},
  booktitle = {IEEE Conference on Computer Vision and Pattern Recognition (CVPR)},
  year      = {2007}
}

@inproceedings{droste2020unisal,
  title     = {Unified image and video saliency modeling},
  author    = {Droste, Richard and Jiao, Jianbo and Noble, J. Alison},
  booktitle = {European Conference on Computer Vision (ECCV)},
  pages     = {419--435},
  year      = {2020}
}

@inproceedings{linardos2021deepgaze,
  title     = {{DeepGaze IIE}: Calibrated prediction in and out-of-domain for state-of-the-art saliency modeling},
  author    = {Linardos, Akis and K{\"u}mmerer, Matthias and Press, Ori and Bethge, Matthias},
  booktitle = {IEEE/CVF International Conference on Computer Vision (ICCV)},
  pages     = {12919--12928},
  year      = {2021}
}

@article{lou2022transalnet,
  title   = {{TranSalNet}: Towards perceptually relevant visual saliency prediction},
  author  = {Lou, Jianxun and Lin, Hanhe and Marshall, David and Saupe, Dietmar and Liu, Hantao},
  journal = {Neurocomputing},
  volume  = {494},
  pages   = {455--467},
  year    = {2022}
}

@article{kummerer2022deepgaze3,
  title   = {{DeepGaze III}: Modeling free-viewing human scanpaths with deep learning},
  author  = {K{\"u}mmerer, Matthias and Bethge, Matthias and Wallis, Thomas S. A.},
  journal = {Journal of Vision},
  volume  = {22},
  number  = {5},
  pages   = {7},
  year    = {2022}
}

@article{dewhurst2012multimatch,
  title   = {It depends on how you look at it: Scanpath comparison in multiple dimensions with {MultiMatch}, a vector-based approach},
  author  = {Dewhurst, Richard and Nystr{\"o}m, Marcus and Jarodzka, Halszka and Foulsham, Tom and Johansson, Roger and Holmqvist, Kenneth},
  journal = {Behavior Research Methods},
  volume  = {44},
  number  = {4},
  pages   = {1079--1100},
  year    = {2012}
}

@article{cristino2010scanmatch,
  title   = {{ScanMatch}: A novel method for comparing fixation sequences},
  author  = {Cristino, Filipe and Math{\^o}t, Sebastiaan and Theeuwes, Jan and Gilchrist, Iain D.},
  journal = {Behavior Research Methods},
  volume  = {42},
  number  = {3},
  pages   = {692--700},
  year    = {2010}
}

@article{mannan1995,
  title   = {Automatic control of saccadic eye movements made in visual inspection of briefly presented 2-{D} images},
  author  = {Mannan, Sabira and Ruddock, Keith H. and Wooding, David S.},
  journal = {Spatial Vision},
  volume  = {9},
  number  = {3},
  pages   = {363--386},
  year    = {1995}
}

@article{pannasch2008ambient,
  title   = {Visual fixation durations and saccade amplitudes: Shifting relationship in a variety of conditions},
  author  = {Pannasch, Sebastian and Helmert, Jens R. and Roth, Katharina and Herbold, Ann-Katrin and Walter, Henrik},
  journal = {Journal of Eye Movement Research},
  volume  = {2},
  number  = {2},
  year    = {2008}
}

@inproceedings{cerf2008faces,
  title     = {Predicting human gaze using low-level saliency combined with face detection},
  author    = {Cerf, Moran and Harel, Jonathan and Einh{\"a}user, Wolfgang and Koch, Christof},
  booktitle = {Advances in Neural Information Processing Systems (NIPS)},
  volume    = {20},
  year      = {2008}
}

@inproceedings{hasler2003colourfulness,
  title     = {Measuring colourfulness in natural images},
  author    = {Hasler, David and S{\"u}sstrunk, Sabine},
  booktitle = {Human Vision and Electronic Imaging VIII, Proc. SPIE},
  volume    = {5007},
  pages     = {87--95},
  year      = {2003}
}

@misc{itu2023p910,
  title        = {Subjective video quality assessment methods for multimedia applications},
  author       = {{ITU-T}},
  howpublished = {Recommendation ITU-T P.910},
  year         = {2023}
}

@book{hambidge1926,
  title     = {The Elements of Dynamic Symmetry},
  author    = {Hambidge, Jay},
  publisher = {Brentano's},
  address   = {New York},
  year      = {1926}
}

@article{otsu1979,
  title   = {A threshold selection method from gray-level histograms},
  author  = {Otsu, Nobuyuki},
  journal = {IEEE Transactions on Systems, Man, and Cybernetics},
  volume  = {9},
  number  = {1},
  pages   = {62--66},
  year    = {1979}
}

@article{benjamini1995fdr,
  title   = {Controlling the false discovery rate: a practical and powerful approach to multiple testing},
  author  = {Benjamini, Yoav and Hochberg, Yosef},
  journal = {Journal of the Royal Statistical Society: Series B},
  volume  = {57},
  number  = {1},
  pages   = {289--300},
  year    = {1995}
}

@article{saleh2015wikiart,
  title   = {Large-scale classification of fine-art paintings: Learning the right metric on the right feature},
  author  = {Saleh, Babak and Elgammal, Ahmed},
  journal = {arXiv preprint arXiv:1505.00855},
  year    = {2015}
}

@book{ishihara1967,
  title     = {Tests for Colour-Blindness},
  author    = {Ishihara, Shinobu},
  publisher = {Kanehara Shuppan},
  year      = {1967}
}

@article{zheng2024birefnet,
  title   = {Bilateral reference for high-resolution dichotomous image segmentation},
  author  = {Zheng, Peng and Gao, Dehong and Fan, Deng-Ping and Liu, Li and Laaksonen, Jorma and Ouyang, Wanli and Sebe, Nicu},
  journal = {CAAI Artificial Intelligence Research},
  volume  = {3},
  pages   = {9150038},
  doi     = {10.26599/AIR.2024.9150038},
  year    = {2024}
}

@misc{ultralytics,
  title        = {Ultralytics {YOLO}, version 8.4 ({YOLO26} and {YOLOE-26} models)},
  author       = {Jocher, Glenn and Qiu, Jing},
  year         = {2026},
  howpublished = {\url{https://github.com/ultralytics/ultralytics}}
}

@inproceedings{wang2025yoloe,
  title     = {{YOLOE}: Real-time seeing anything},
  author    = {Wang, Ao and Liu, Lihao and Chen, Hui and Lin, Zijia and Han, Jungong and Ding, Guiguang},
  booktitle = {IEEE/CVF International Conference on Computer Vision (ICCV)},
  year      = {2025}
}

@inproceedings{hong2021cacnet,
  title     = {Composing photos like a photographer},
  author    = {Hong, Chaoyi and Du, Shuaiyuan and Xian, Ke and Lu, Hao and Cao, Zhiguo and Zhong, Weicai},
  booktitle = {IEEE/CVF Conference on Computer Vision and Pattern Recognition (CVPR)},
  pages     = {7057--7066},
  year      = {2021}
}

@article{lee2018kupcp,
  title   = {Photographic composition classification and dominant geometric element detection for outdoor scenes},
  author  = {Lee, Jun-Tae and Kim, Han-Ul and Lee, Chul and Kim, Chang-Su},
  journal = {Journal of Visual Communication and Image Representation},
  volume  = {55},
  pages   = {91--105},
  year    = {2018}
}

@inproceedings{radford2021clip,
  title     = {Learning transferable visual models from natural language supervision},
  author    = {Radford, Alec and Kim, Jong Wook and Hallacy, Chris and Ramesh, Aditya and Goh, Gabriel and Agarwal, Sandhini and Sastry, Girish and Askell, Amanda and Mishkin, Pamela and Clark, Jack and Krueger, Gretchen and Sutskever, Ilya},
  booktitle = {International Conference on Machine Learning (ICML)},
  pages     = {8748--8763},
  year      = {2021}
}

@article{oquab2024dinov2,
  title   = {{DINOv2}: Learning robust visual features without supervision},
  author  = {Oquab, Maxime and Darcet, Timoth{\'e}e and Moutakanni, Th{\'e}o and others},
  journal = {Transactions on Machine Learning Research},
  year    = {2024}
}

@article{simeoni2025dinov3,
  title   = {{DINOv3}},
  author  = {Sim{\'e}oni, Oriane and Vo, Huy V. and Seitzer, Maximilian and others},
  journal = {arXiv preprint arXiv:2508.10104},
  year    = {2025}
}

@inproceedings{assran2023ijepa,
  title     = {Self-supervised learning from images with a joint-embedding predictive architecture},
  author    = {Assran, Mahmoud and Duval, Quentin and Misra, Ishan and Bojanowski, Piotr and Vincent, Pascal and Rabbat, Michael and LeCun, Yann and Ballas, Nicolas},
  booktitle = {IEEE/CVF Conference on Computer Vision and Pattern Recognition (CVPR)},
  pages     = {15619--15629},
  year      = {2023}
}

@inproceedings{bardes2022vicreg,
  title     = {{VICReg}: Variance-invariance-covariance regularization for self-supervised learning},
  author    = {Bardes, Adrien and Ponce, Jean and LeCun, Yann},
  booktitle = {International Conference on Learning Representations (ICLR)},
  year      = {2022}
}

@inproceedings{he2022mae,
  title     = {Masked autoencoders are scalable vision learners},
  author    = {He, Kaiming and Chen, Xinlei and Xie, Saining and Li, Yanghao and Doll{\'a}r, Piotr and Girshick, Ross},
  booktitle = {IEEE/CVF Conference on Computer Vision and Pattern Recognition (CVPR)},
  pages     = {16000--16009},
  year      = {2022}
}

@inproceedings{dosovitskiy2021vit,
  title     = {An image is worth 16x16 words: Transformers for image recognition at scale},
  author    = {Dosovitskiy, Alexey and Beyer, Lucas and Kolesnikov, Alexander and Weissenborn, Dirk and Zhai, Xiaohua and Unterthiner, Thomas and Dehghani, Mostafa and Minderer, Matthias and Heigold, Georg and Gelly, Sylvain and Uszkoreit, Jakob and Houlsby, Neil},
  booktitle = {International Conference on Learning Representations (ICLR)},
  year      = {2021}
}

@inproceedings{he2016resnet,
  title     = {Deep residual learning for image recognition},
  author    = {He, Kaiming and Zhang, Xiangyu and Ren, Shaoqing and Sun, Jian},
  booktitle = {IEEE Conference on Computer Vision and Pattern Recognition (CVPR)},
  pages     = {770--778},
  year      = {2016}
}

@inproceedings{simonyan2015vgg,
  title     = {Very deep convolutional networks for large-scale image recognition},
  author    = {Simonyan, Karen and Zisserman, Andrew},
  booktitle = {International Conference on Learning Representations (ICLR)},
  year      = {2015}
}

@inproceedings{liu2022convnext,
  title     = {A {ConvNet} for the 2020s},
  author    = {Liu, Zhuang and Mao, Hanzi and Wu, Chao-Yuan and Feichtenhofer, Christoph and Darrell, Trevor and Xie, Saining},
  booktitle = {IEEE/CVF Conference on Computer Vision and Pattern Recognition (CVPR)},
  pages     = {11976--11986},
  year      = {2022}
}

@inproceedings{zhang2020bertscore,
  title     = {{BERTScore}: Evaluating text generation with {BERT}},
  author    = {Zhang, Tianyi and Kishore, Varsha and Wu, Felix and Weinberger, Kilian Q. and Artzi, Yoav},
  booktitle = {International Conference on Learning Representations (ICLR)},
  year      = {2020}
}

@misc{helsinki2013,
  title        = {{WMA} Declaration of {Helsinki} -- Ethical principles for medical research involving human subjects},
  author       = {{World Medical Association}},
  year         = {2013},
  howpublished = {\url{https://www.wma.net/policies-post/wma-declaration-of-helsinki/}}
}

@phdthesis{kerkouri2024thesis,
  title  = {A Gaze into the Art World: Predicting Visual Attention Using Deep Learning},
  author = {Kerkouri, Mohamed Amine},
  school = {Universit{\'e} d'Orl{\'e}ans},
  year   = {2024}
}

@inproceedings{kerkouri2021salypath,
  title     = {{SalyPath}: A deep-based architecture for visual attention prediction},
  author    = {Kerkouri, Mohamed Amine and Tliba, Marouane and Chetouani, Aladine and Harba, Rachid},
  booktitle = {IEEE International Conference on Image Processing (ICIP)},
  pages     = {1464--1468},
  year      = {2021},
  doi       = {10.1109/ICIP42928.2021.9506295}
}

@article{tliba2022satsal,
  title   = {{SATSal}: A multi-level self-attention based architecture for visual saliency prediction},
  author  = {Tliba, Marouane and Kerkouri, Mohamed Amine and Ghariba, Bashir and Chetouani, Aladine and {\c{C}}{\"o}ltekin, Arzu and Shehata, Mohamed Sami and Bruno, Alessandro},
  journal = {IEEE Access},
  volume  = {10},
  pages   = {20701--20713},
  year    = {2022},
  doi     = {10.1109/ACCESS.2022.3152189}
}

@article{kerkouri2022salypath360,
  title   = {{SalyPath360}: Saliency and scanpath prediction framework for omnidirectional images},
  author  = {Kerkouri, Mohamed Amine and Tliba, Marouane and Chetouani, Aladine and Sayeh, Mohamed},
  journal = {Electronic Imaging},
  volume  = {34},
  number  = {11},
  pages   = {168-1--168-7},
  year    = {2022},
  doi     = {10.2352/EI.2022.34.11.HVEI-168}
}

@article{kerkouri2022simple,
  title   = {A simple and efficient deep scanpath prediction},
  author  = {Kerkouri, Mohamed Amine and Chetouani, Aladine},
  journal = {Electronic Imaging},
  volume  = {34},
  number  = {11},
  pages   = {188-1--188-7},
  year    = {2022},
  doi     = {10.2352/EI.2022.34.11.HVEI-188}
}

@inproceedings{kerkouri2022domain,
  title     = {A domain adaptive deep learning solution for scanpath prediction of paintings},
  author    = {Kerkouri, Mohamed Amine and Tliba, Marouane and Chetouani, Aladine and Bruno, Alessandro},
  booktitle = {International Conference on Content-Based Multimedia Indexing (CBMI)},
  pages     = {57--63},
  year      = {2022},
  doi       = {10.1145/3549555.3549597}
}

@inproceedings{tliba2022selfsupervised,
  title     = {Self supervised scanpath prediction framework for painting images},
  author    = {Tliba, Marouane and Kerkouri, Mohamed Amine and Chetouani, Aladine and Bruno, Alessandro},
  booktitle = {IEEE/CVF Conference on Computer Vision and Pattern Recognition Workshops (CVPRW)},
  pages     = {1538--1547},
  year      = {2022},
  doi       = {10.1109/CVPRW56347.2022.00160}
}

@inproceedings{kerkouri2023adversarial,
  title     = {An inter-observer consistent deep adversarial training for visual scanpath prediction},
  author    = {Kerkouri, Mohamed Amine and Tliba, Marouane and Chetouani, Aladine and Bruno, Alessandro},
  booktitle = {IEEE International Conference on Image Processing (ICIP)},
  pages     = {2595--2599},
  year      = {2023},
  doi       = {10.1109/ICIP49359.2023.10222686}
}

@article{kerkouri2026spgen,
  title   = {{SPGen}: Stochastic scanpath generation for paintings using unsupervised domain adaptation},
  author  = {Kerkouri, Mohamed Amine and Tliba, Marouane and Chetouani, Aladine and Bruno, Alessandro},
  journal = {arXiv preprint arXiv:2602.22049},
  year    = {2026}
}

@inproceedings{kerkouri2026whatsaw,
  title     = {What They Saw, Not Just Where They Looked: Semantic Scanpath Similarity via {VLMs} and {NLP} Metrics},
  author    = {Kerkouri, Mohamed Amine and Tliba, Marouane and Wang, Bin and Chetouani, Aladine and Bagci, Ulas and Bruno, Alessandro},
  booktitle = {Proceedings of the 2026 Symposium on Eye Tracking Research and Applications (ETRA '26)},
  pages     = {1--7},
  year      = {2026},
  publisher = {ACM},
  doi       = {10.1145/3797246.3806223}
}

@inproceedings{bruno2023cvd,
  title     = {Detecting colour vision deficiencies via webcam-based eye-tracking: A case study},
  author    = {Bruno, Alessandro and Tliba, Marouane and Kerkouri, Mohamed Amine and Chetouani, Aladine and Giunta, Carlo Calogero and {\c{C}}{\"o}ltekin, Arzu},
  booktitle = {Proceedings of the 2023 Symposium on Eye Tracking Research and Applications (ETRA '23)},
  pages     = {1--2},
  year      = {2023},
  publisher = {ACM},
  doi       = {10.1145/3588015.3590133}
}

@inproceedings{tliba2024mae,
  title     = {Perceptual evaluation of masked autoencoder emergent properties through eye-tracking-based policy},
  author    = {Tliba, Marouane and Kerkouri, Mohamed Amine and Chetouani, Aladine and Bruno, Alessandro and El Hassouni, Mohammed and {\c{C}}{\"o}ltekin, Arzu},
  booktitle = {Proceedings of the 2024 Symposium on Eye Tracking Research and Applications (ETRA '24)},
  pages     = {1--3},
  year      = {2024},
  publisher = {ACM},
  doi       = {10.1145/3649902.3655638}
}

@inproceedings{kerkouri2024quantization,
  title     = {Quantization effects on neural networks perception: How would quantization change the perceptual field of vision models?},
  author    = {Kerkouri, Mohamed Amine and Tliba, Marouane and Chetouani, Aladine and Bruno, Alessandro},
  booktitle = {IEEE International Conference on Image Processing Theory, Tools and Applications (IPTA)},
  pages     = {1--6},
  year      = {2024},
  doi       = {10.1109/IPTA62886.2024.10755934}
}

@inproceedings{wong2025shifts,
  title     = {Shifts in doctors' eye movements between real and {AI}-generated medical images},
  author    = {Wong, David C. and Wang, Bin and Durak, Gorkem and Tliba, Marouane and Kerkouri, Mohamed Amine and Chetouani, Aladine and Cetin, Ahmet Enis and Topel, Cagdas and others},
  booktitle = {Proceedings of the 2025 Symposium on Eye Tracking Research and Applications (ETRA '25)},
  pages     = {1--7},
  year      = {2025},
  publisher = {ACM},
  doi       = {10.1145/3715669.3726789}
}

@article{kerkouri2026divergent,
  title   = {Divergent Gaze Patterns in Artistic Viewing: Spatial and Temporal Signatures of Attention Across Autistic Individuals, Artists, and Neurotypical Observers},
  author  = {Kerkouri, Mohamed Amine and Senggaran, Daphn{\'e} and Jusiak, Renaud and Lehmann, Oc{\'e}ane and Tliba, Marouane and Wardak, Claire and Houy-Durand, Emmanuelle and Morel-Kohlmeyer, Shasha and Chetouani, Aladine and Aguillon-Hernandez, Nadia},
  journal = {arXiv preprint arXiv:2607.15227},
  year    = {2026}
}

@article{kerkouri2026foveated,
  title   = {Matched Outcomes, Divergent Gaze: How Foveated {MLLMs} Search Compared to Humans},
  author  = {Kerkouri, Mohamed Amine and Tliba, Marouane and Chetouani, Aladine and Bagci, Ulas and Bruno, Alessandro},
  journal = {arXiv preprint arXiv:2608.16514},
  year    = {2026}
}

@article{benyahia2026gazerefine,
  title   = {{GazeRefine}: Expert Gaze as a Test-Time Prompt for Training-Free Medical Image Segmentation},
  author  = {Benyahia, Mohammed Oussama and Tliba, Marouane and Kerkouri, Mohamed Amine and Yousra, Taifour and Wang, Bin and Bengtsson, Max and Durak, Gorkem and Keles, Elif and Ming, Zuheng and Penhaker, Marek and others},
  journal = {arXiv preprint arXiv:2609.01310},
  year    = {2026}
}
}

\clearpage
\setcounter{page}{1}
\onecolumn
\begin{center}
  {\Large\textbf{\avatt: A Free-Viewing Eye-Tracking Dataset and Benchmark for Visual Attention on Paintings}}\\[6pt]
  {\large Supplementary Material}
\end{center}
\appendix
\renewcommand{\thefigure}{S\arabic{figure}}
\renewcommand{\thetable}{S\arabic{table}}
\setcounter{figure}{0}
\setcounter{table}{0}

\section{Participants and protocol}

\begin{figure}[h]
  \centering
  \includegraphics[width=\textwidth]{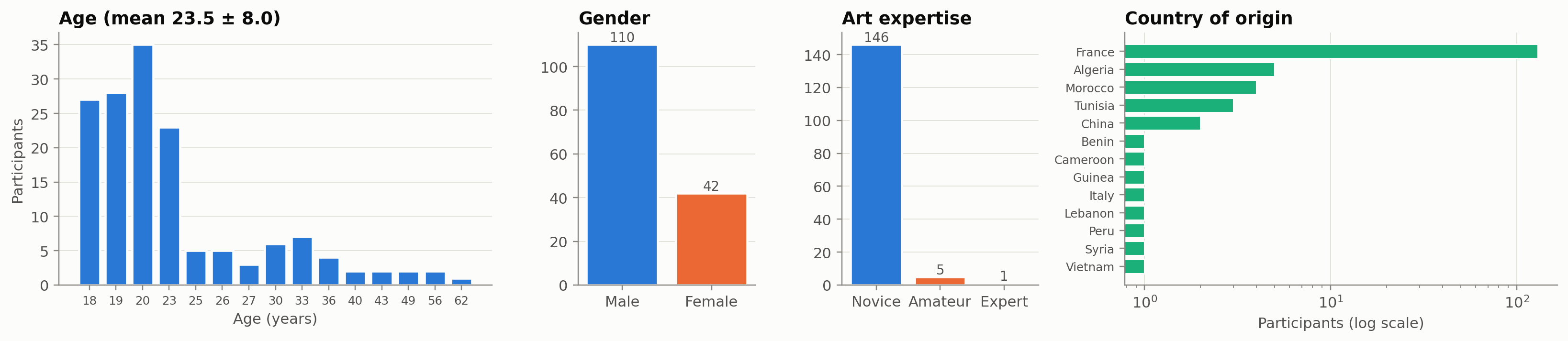}
  \caption{Participants: age, gender, self-reported art expertise and country of origin.}
\end{figure}

The participants were students, researchers and staff, recruited by convenience sampling, with no compensation. Colour
vision was screened with three Ishihara plates (12, 72, 42); a few participants showed mild red--green deficiencies,
without visible effects on their gaze. The eye tracker was positioned below the screen at 50--70\,cm from the
participant. \distPctRange\% of fixations were made at 50--65\,cm.

\section{Additional dataset statistics}

\begin{figure}[h]
  \centering
  \includegraphics[width=0.9\textwidth]{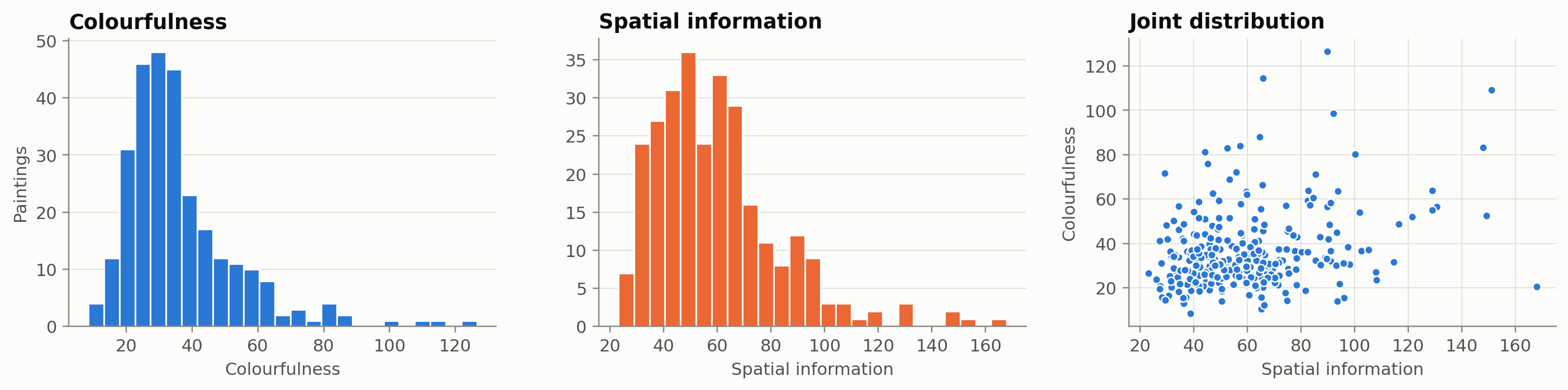}
  \caption{Colourfulness~\cite{hasler2003colourfulness} and spatial information~\cite{itu2023p910} of the paintings
  (painting area only). Mean colourfulness \imgColourMean{} $\pm$ \imgColourSD, SI \imgSIMean{} $\pm$ \imgSISD; Pearson
  $r=\colSIpearson$.}
\end{figure}

\begin{figure}[h]
  \centering
  \includegraphics[width=0.8\textwidth]{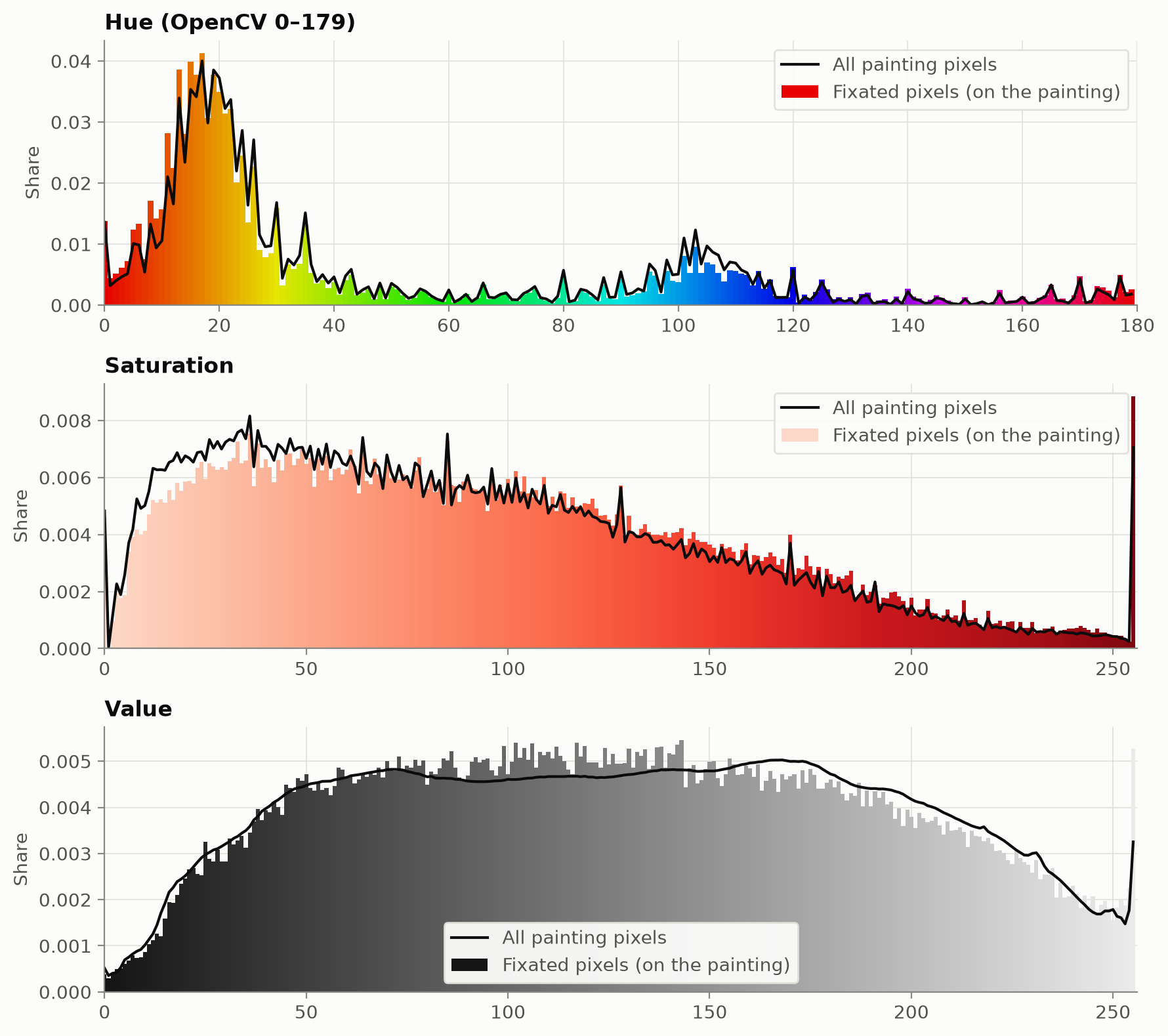}
  \caption{Hue, saturation and value at fixated pixels (bars) vs all painting pixels (line).}
\end{figure}

\begin{figure}[h]
  \centering
  \begin{subfigure}{0.32\textwidth}\includegraphics[width=\linewidth]{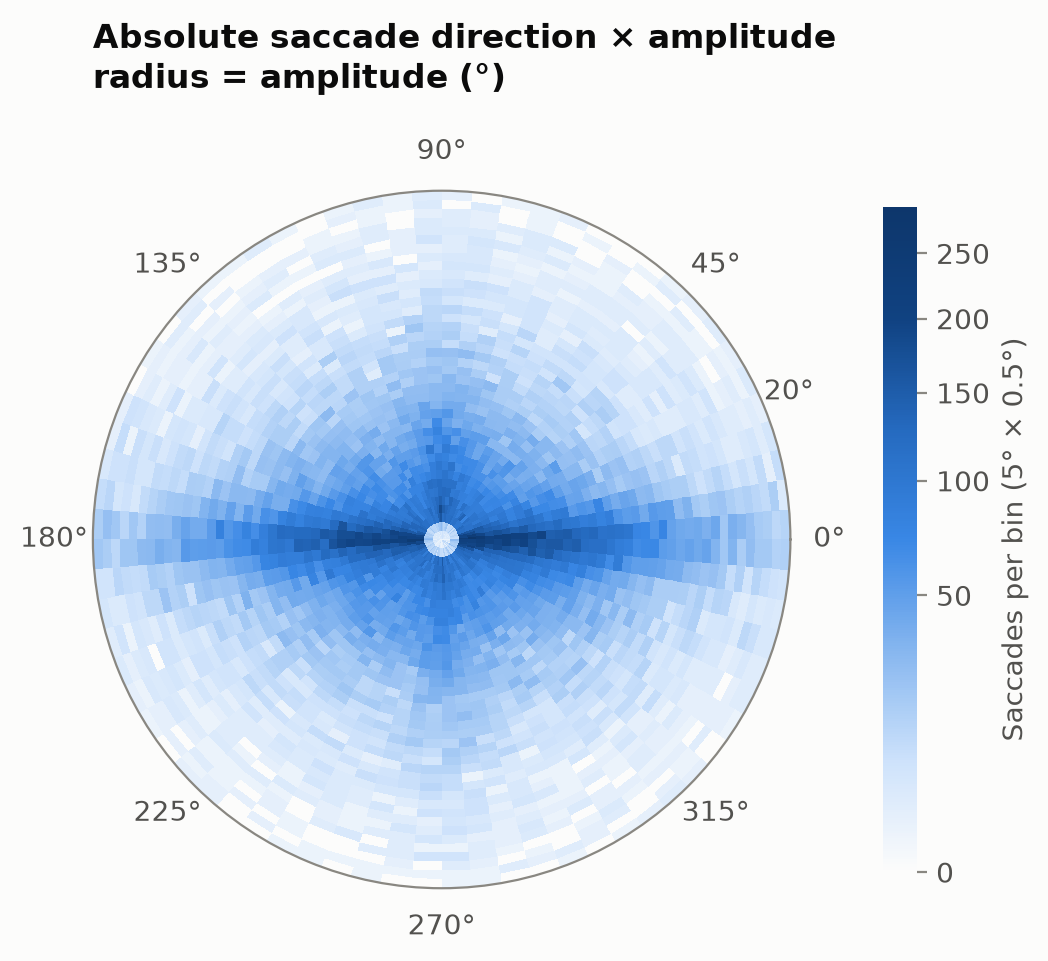}\end{subfigure}
  \begin{subfigure}{0.32\textwidth}\includegraphics[width=\linewidth]{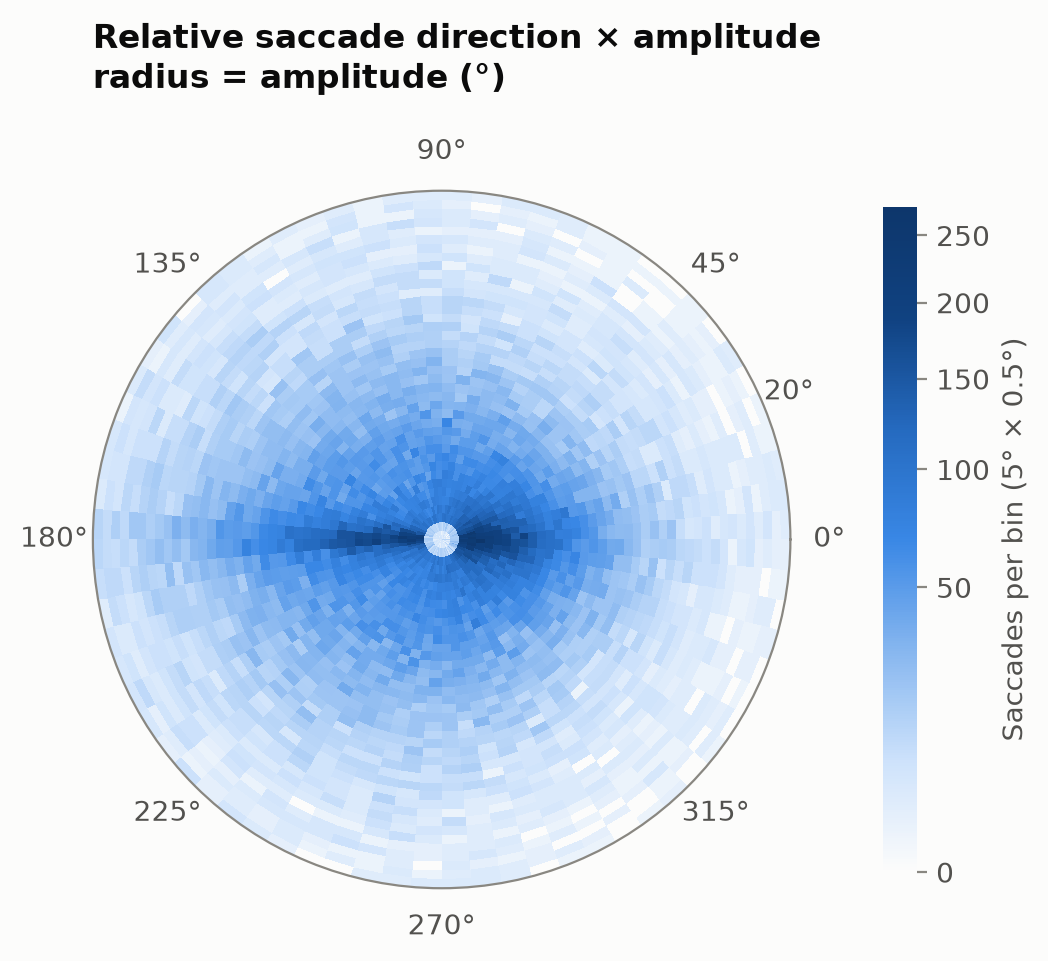}\end{subfigure}
  \begin{subfigure}{0.34\textwidth}\includegraphics[width=\linewidth]{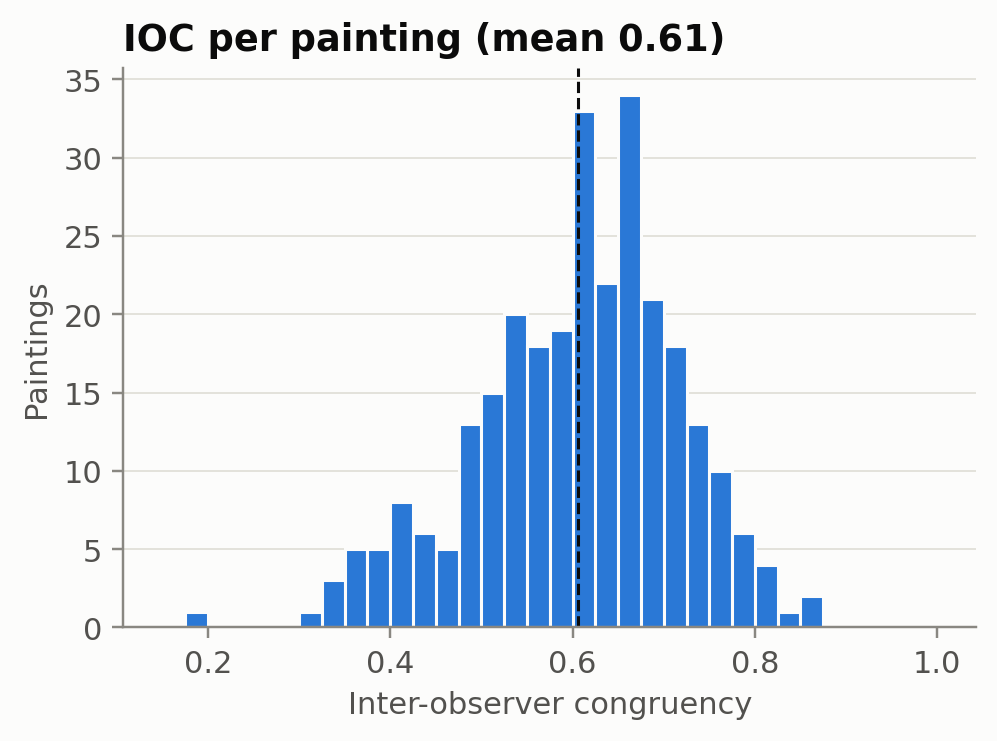}\end{subfigure}
  \caption{Saccade direction $\times$ amplitude (absolute and relative) and inter-observer congruency per painting.}
\end{figure}

\begin{figure}[h]
  \centering
  \includegraphics[width=0.6\textwidth]{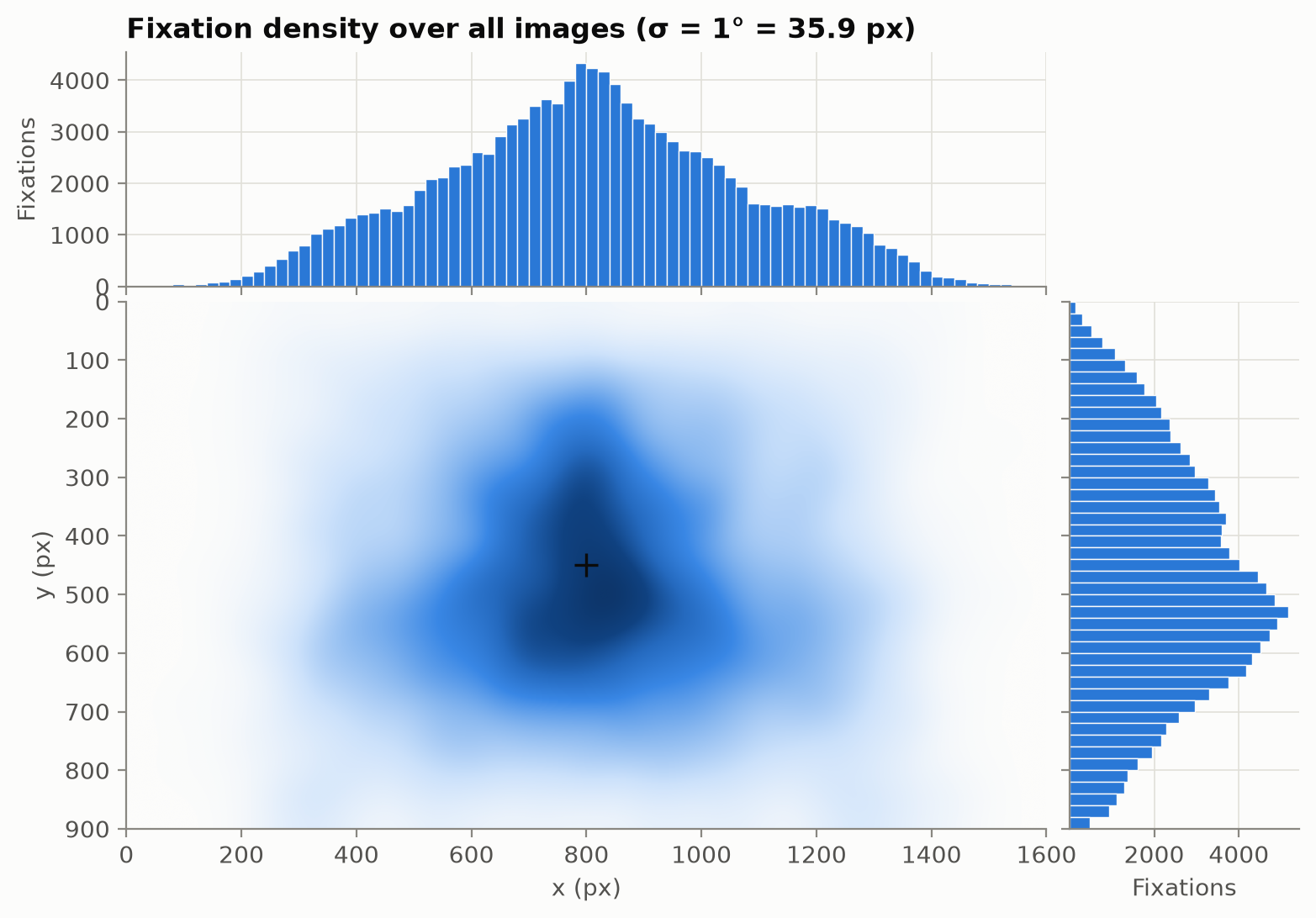}
  \caption{Fixation density over all paintings ($\sigma=1\dg$), with marginals.}
\end{figure}

\begin{figure}[h]
  \centering
  \includegraphics[width=\textwidth]{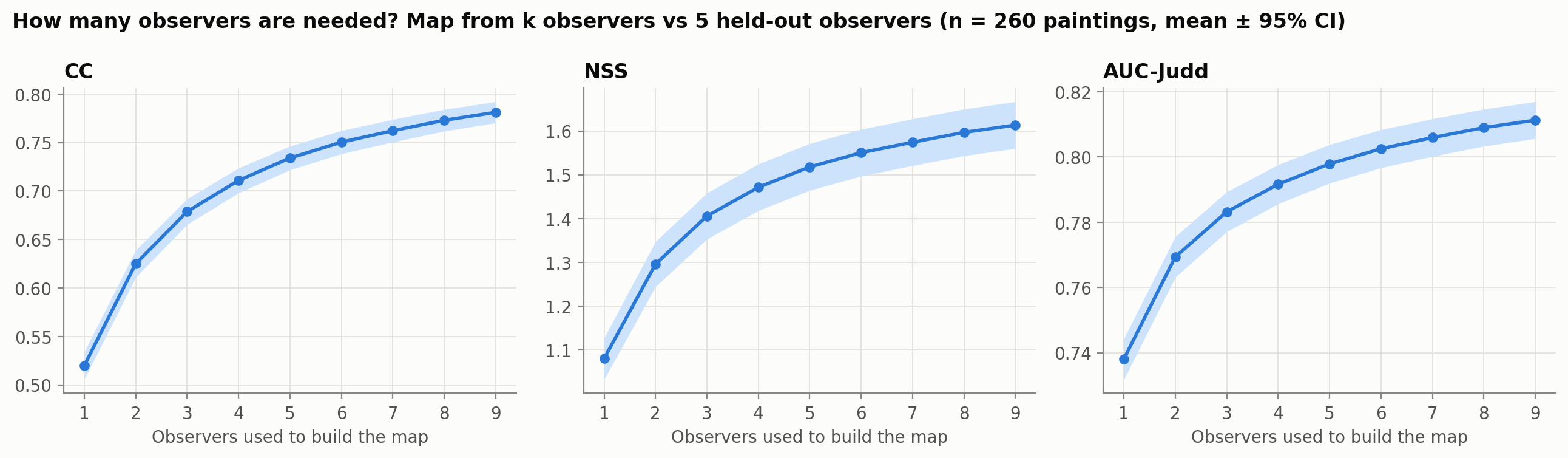}
  \caption{Maps from $k$ observers vs five held-out observers (\convImages{} paintings, mean $\pm$ 95\% CI).}
  \label{fig:supp-convergence}
\end{figure}

\section{Benchmarks}

\begin{figure}[h]
  \centering
  \includegraphics[width=\textwidth]{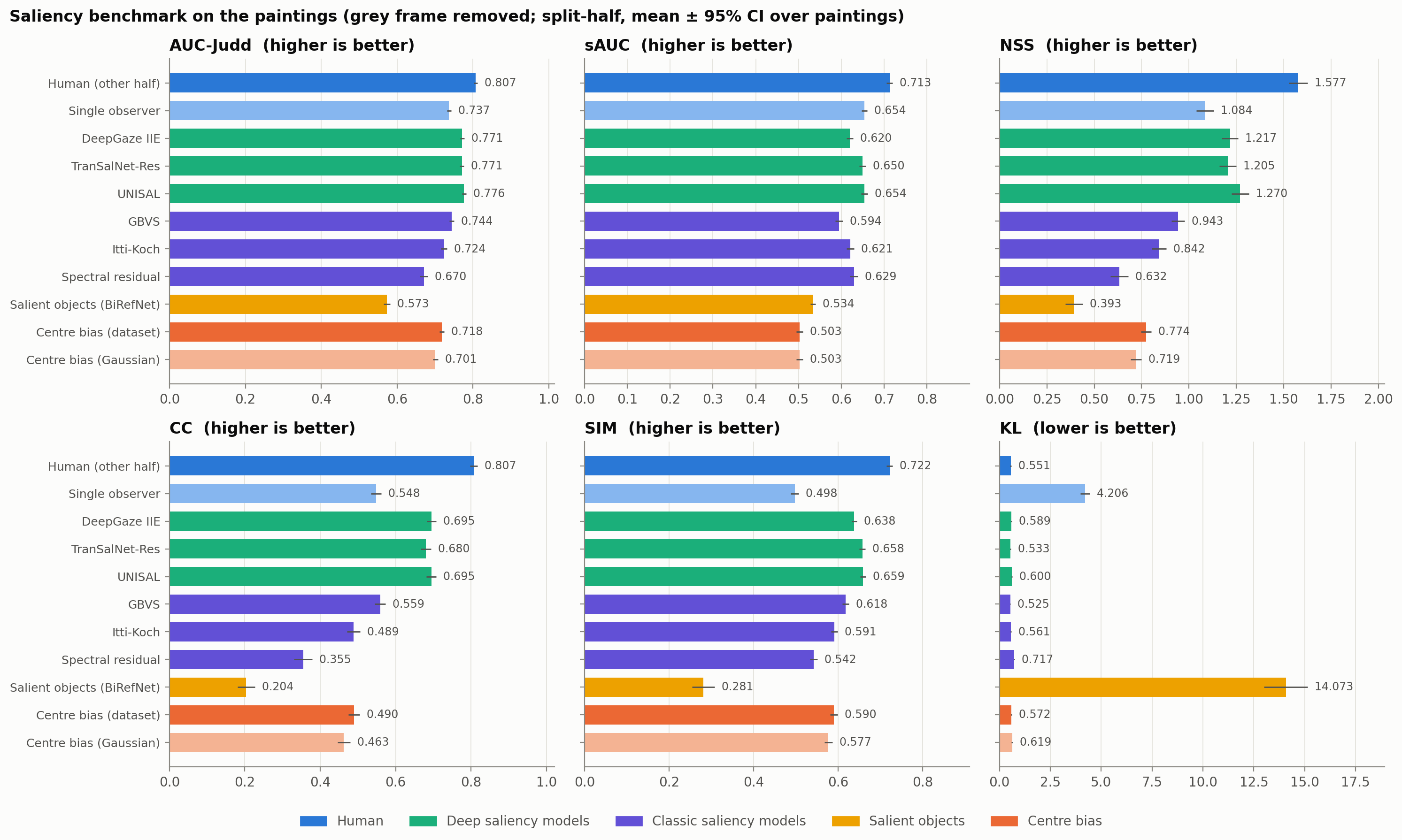}
  \caption{Saliency benchmark on the paintings (mean $\pm$ 95\% CI over paintings).}
\end{figure}

\begin{figure}[h]
  \centering
  \includegraphics[width=\textwidth]{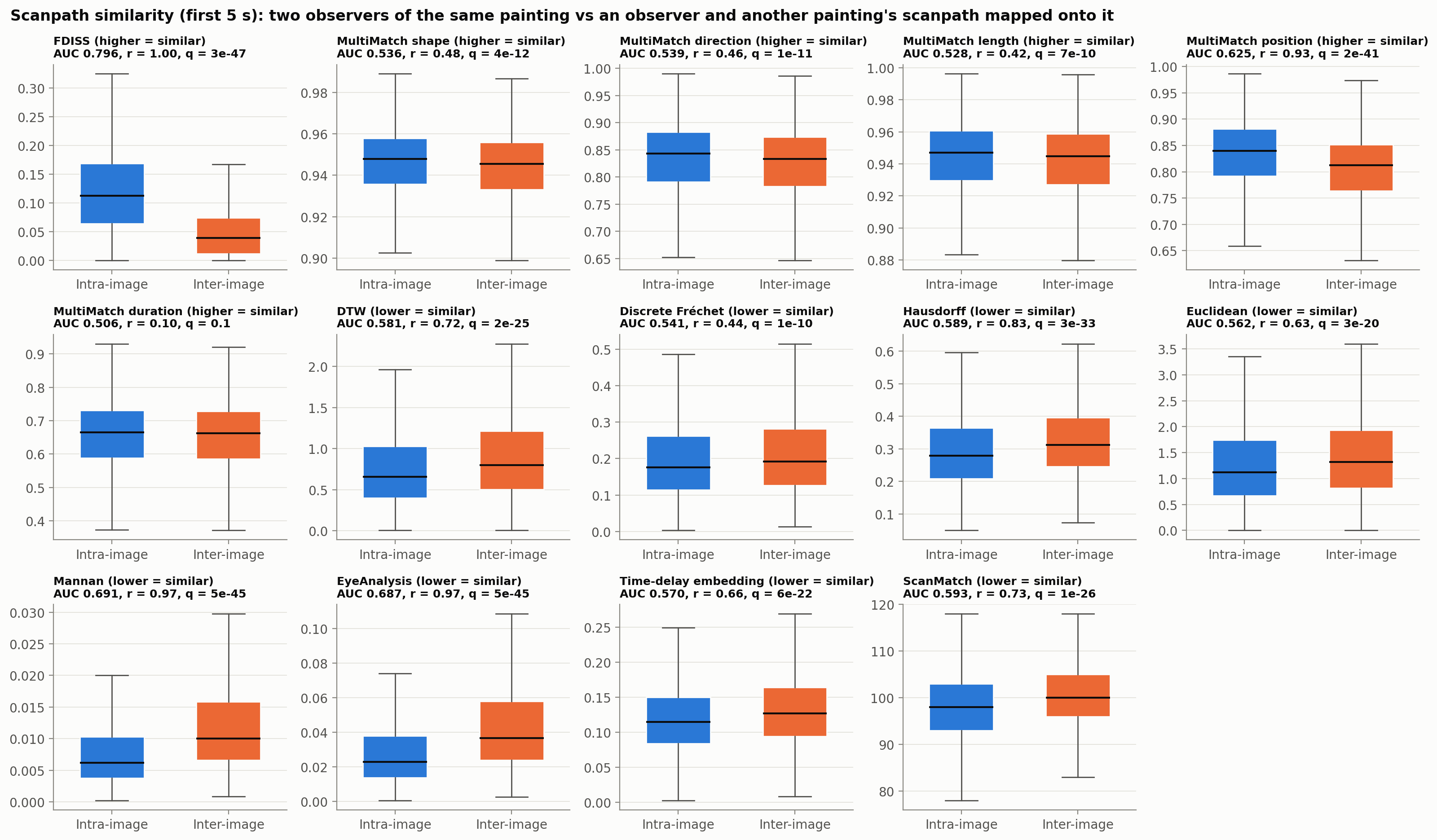}
  \caption{Spatial scanpath measures: intra- vs inter-image pairs.}
\end{figure}

\begin{figure}[h]
  \centering
  \includegraphics[width=0.75\textwidth]{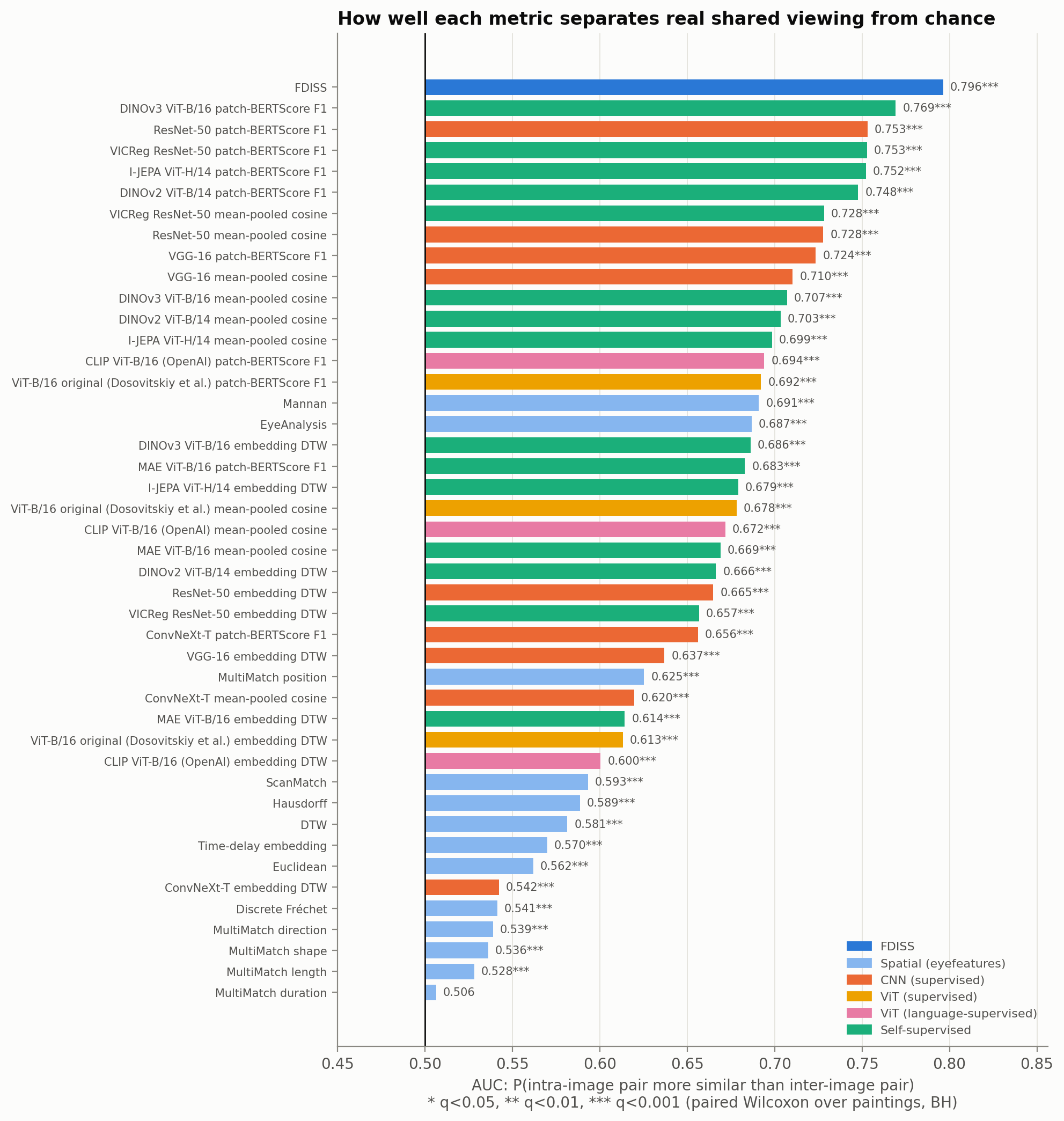}
  \caption{All 44 scanpath measures ranked by AUC (intra vs inter).}
\end{figure}

\section{Content and composition}

\begin{figure}[h]
  \centering
  \includegraphics[width=\textwidth]{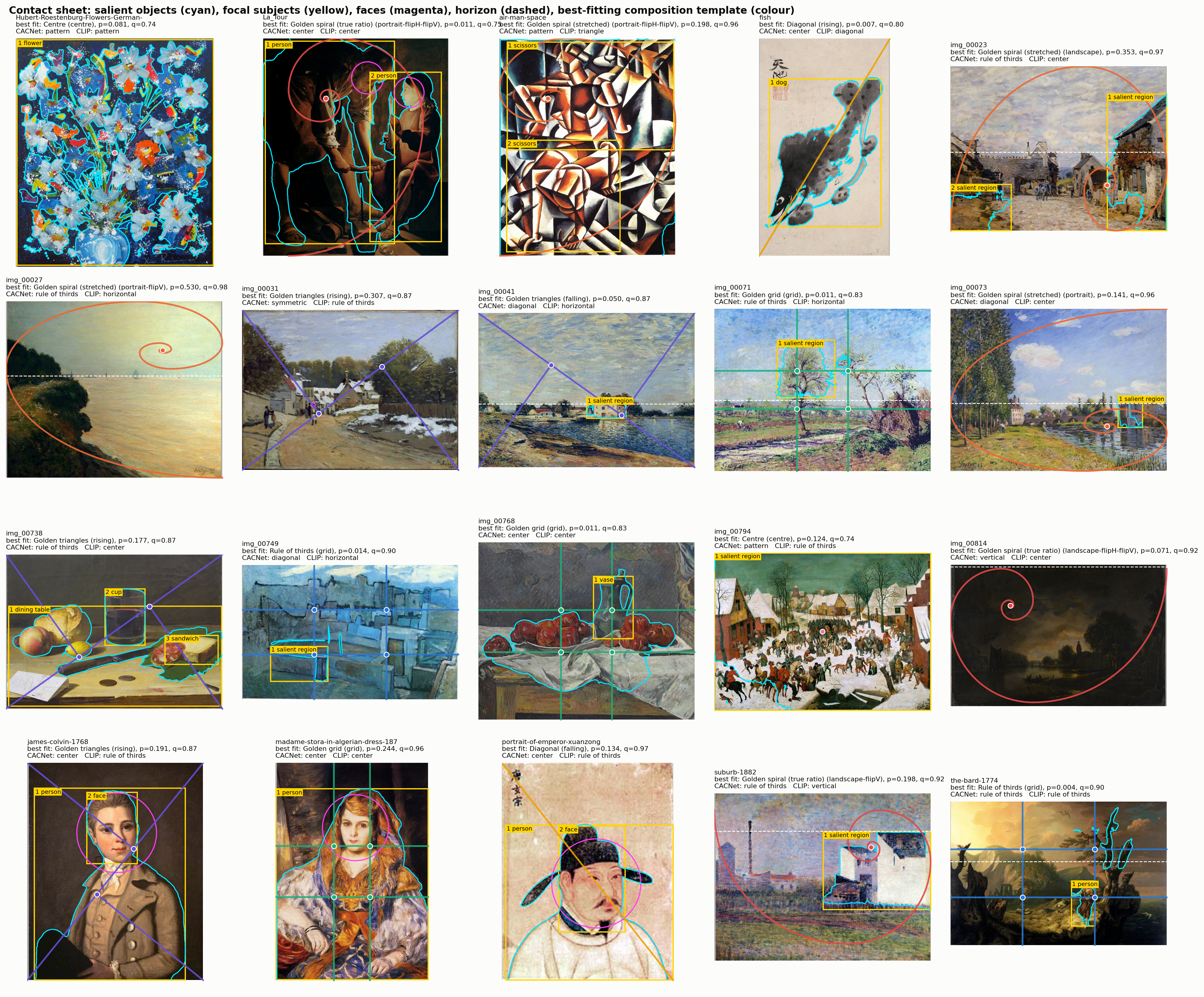}
  \caption{Image-only detections: salient objects (cyan), focal subjects (yellow), faces (magenta), horizon (dashed) and
  best-fitting composition template.}
\end{figure}

\begin{figure}[h]
  \centering
  \includegraphics[width=\textwidth]{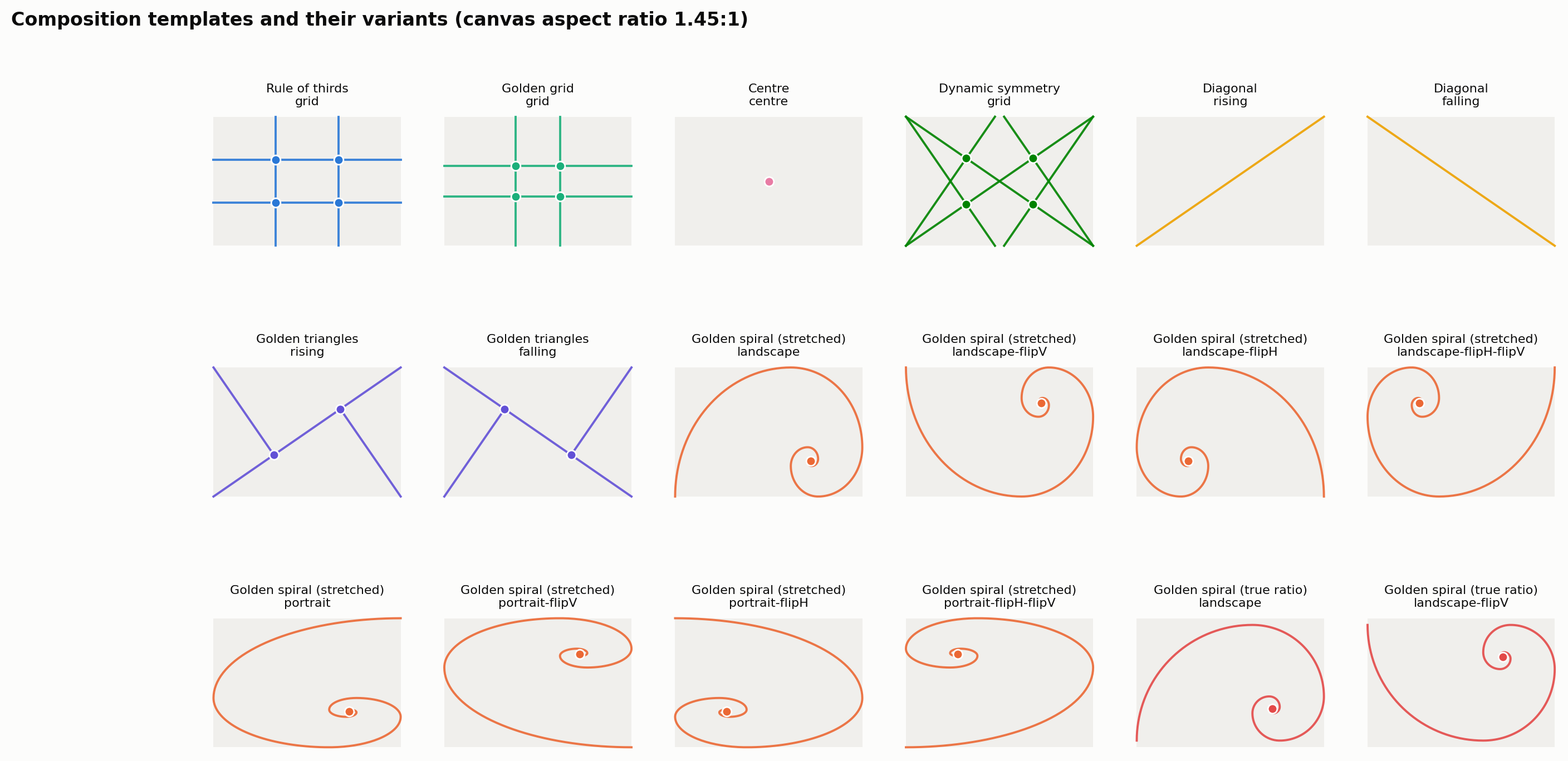}
  \caption{Composition templates and their variants (aspect ratio 1.45).}
  \label{fig:supp-templates}
\end{figure}

\mypara{Fitting details.} The \emph{point fit} of a variant is the salience-weighted mean of
$\exp(-d^2/2\sigma^2)$ over focal subjects, where $d$ is the distance from the subject's centroid to the nearest key
point as a fraction of the painting diagonal and $\sigma=0.05$. The \emph{curve fit} is $\log_2$ of the mean of a
structure map (edge magnitude weighted by salient-object probability, blurred and normalised to mean 1) along the
template's lines. Each component is turned into a percentile against the other paintings' contents on the same
template, the two percentiles are averaged, and the best variant is kept. A per-painting $p$-value is the share of
null paintings whose best-variant statistic is at least as high. This test is deliberately conservative: shared
conventions, such as subjects near the centre or on the thirds, are part of the null, so it measures how much a
painting follows a rule \emph{beyond} paintings in general. We therefore use the fit as a graded score in the main
paper, and release the per-painting fits and $p$-values with the code.

\begin{figure}[h]
  \centering
  \includegraphics[width=0.85\textwidth]{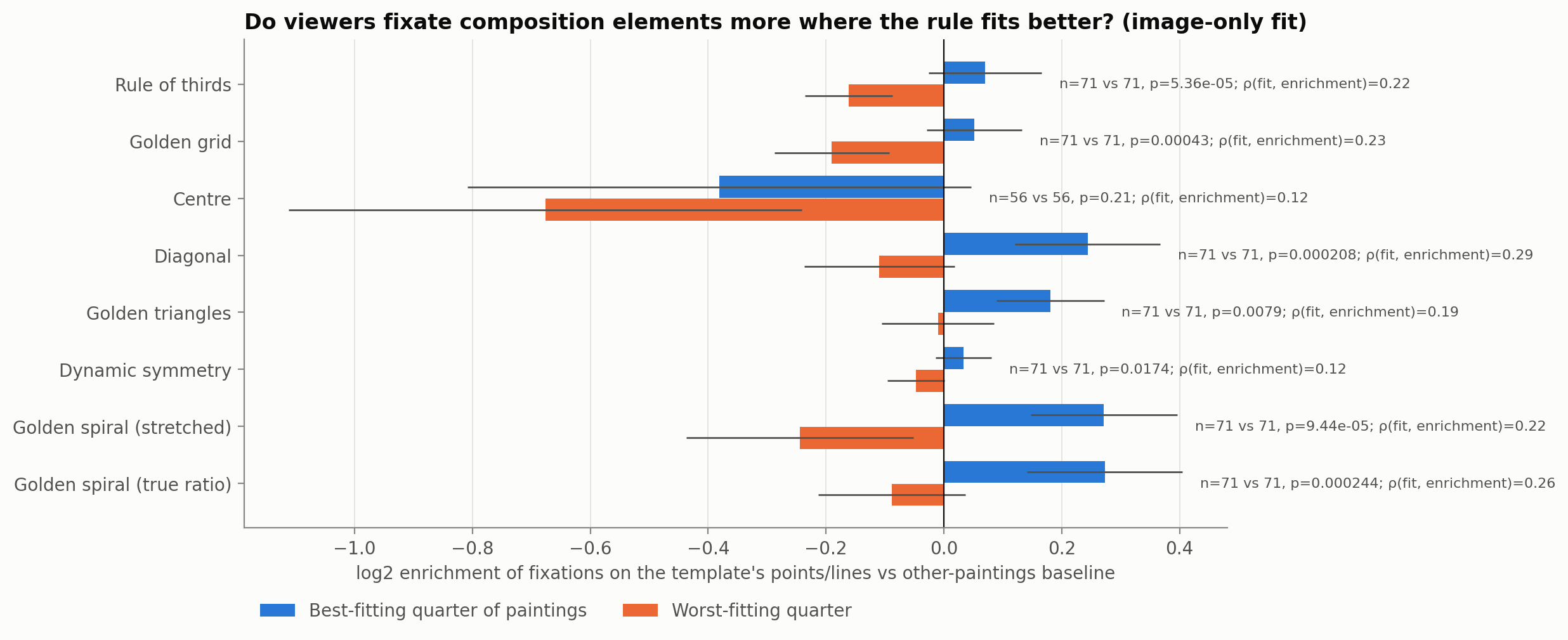}
  \caption{Fixation enrichment on composition templates, for the best- and worst-fitting quarter of paintings.}
\end{figure}

\begin{figure}[h]
  \centering
  \includegraphics[width=\textwidth]{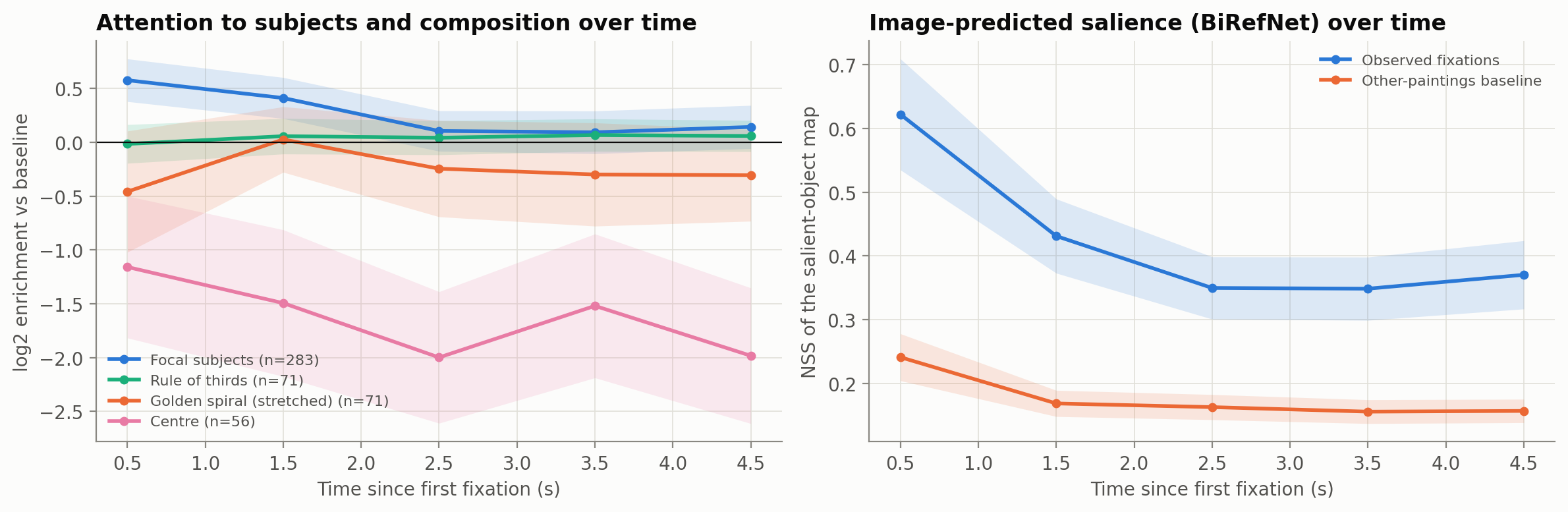}
  \caption{Attention to focal subjects and composition elements over time (left); NSS of the salient-object map over
  time (right).}
\end{figure}

\begin{figure}[h]
  \centering
  \includegraphics[width=0.8\textwidth]{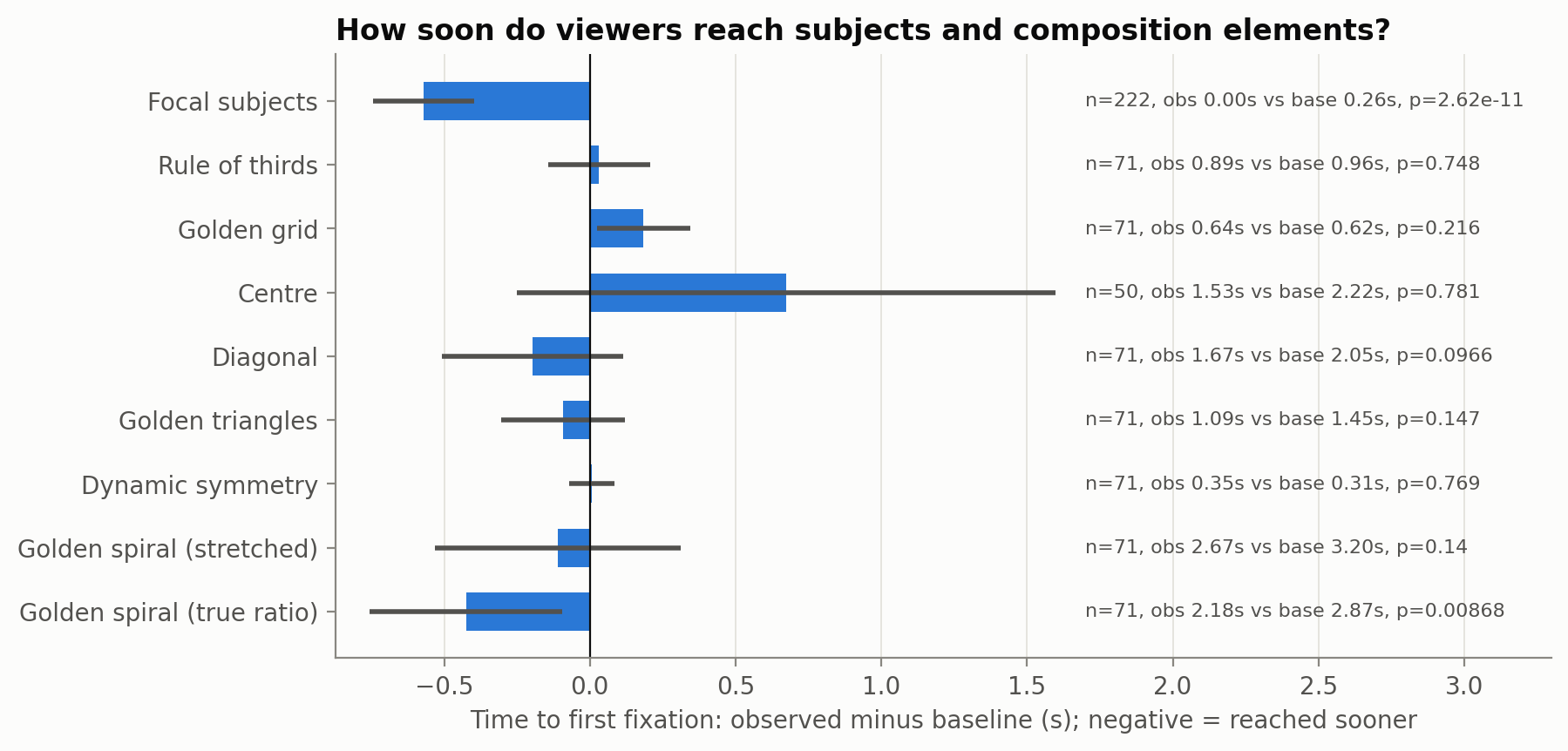}
  \caption{Time to first fixation, observed minus baseline.}
\end{figure}

\begin{figure}[h]
  \centering
  \includegraphics[width=0.8\textwidth]{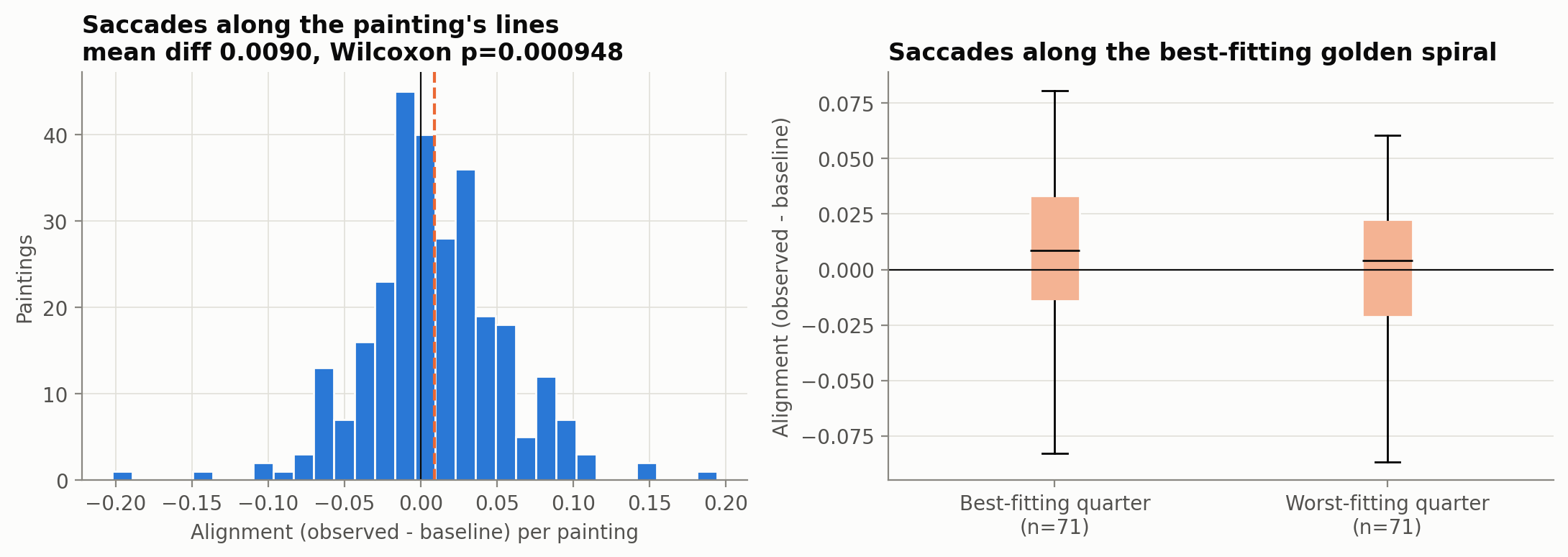}
  \caption{Saccade alignment with the painting's lines and with the best-fitting golden spiral.}
\end{figure}

\begin{figure}[h]
  \centering
  \includegraphics[width=\textwidth]{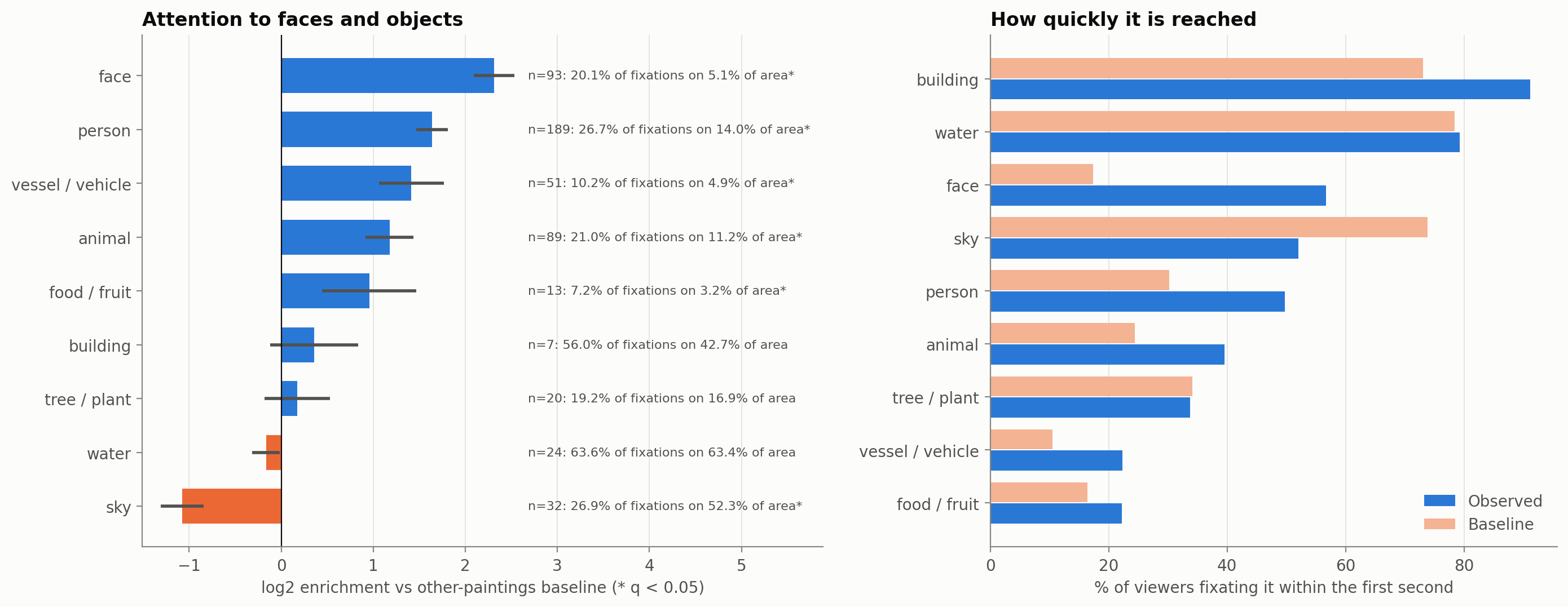}
  \caption{Attention to faces and object categories.}
\end{figure}

\begin{figure}[h]
  \centering
  \includegraphics[width=0.7\textwidth]{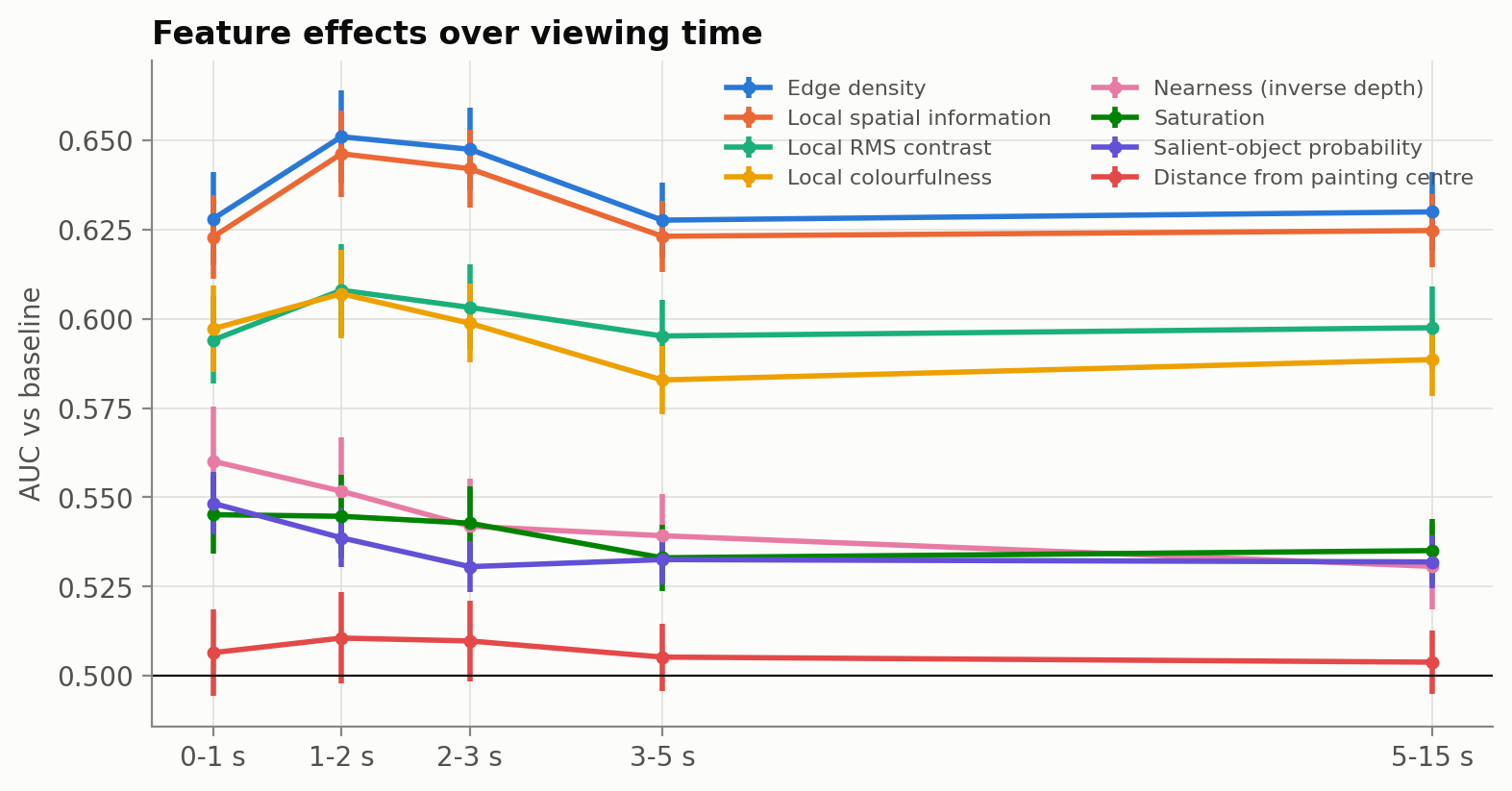}
  \caption{Image-feature effects over viewing time.}
\end{figure}

\begin{figure}[h]
  \centering
  \includegraphics[width=0.75\textwidth]{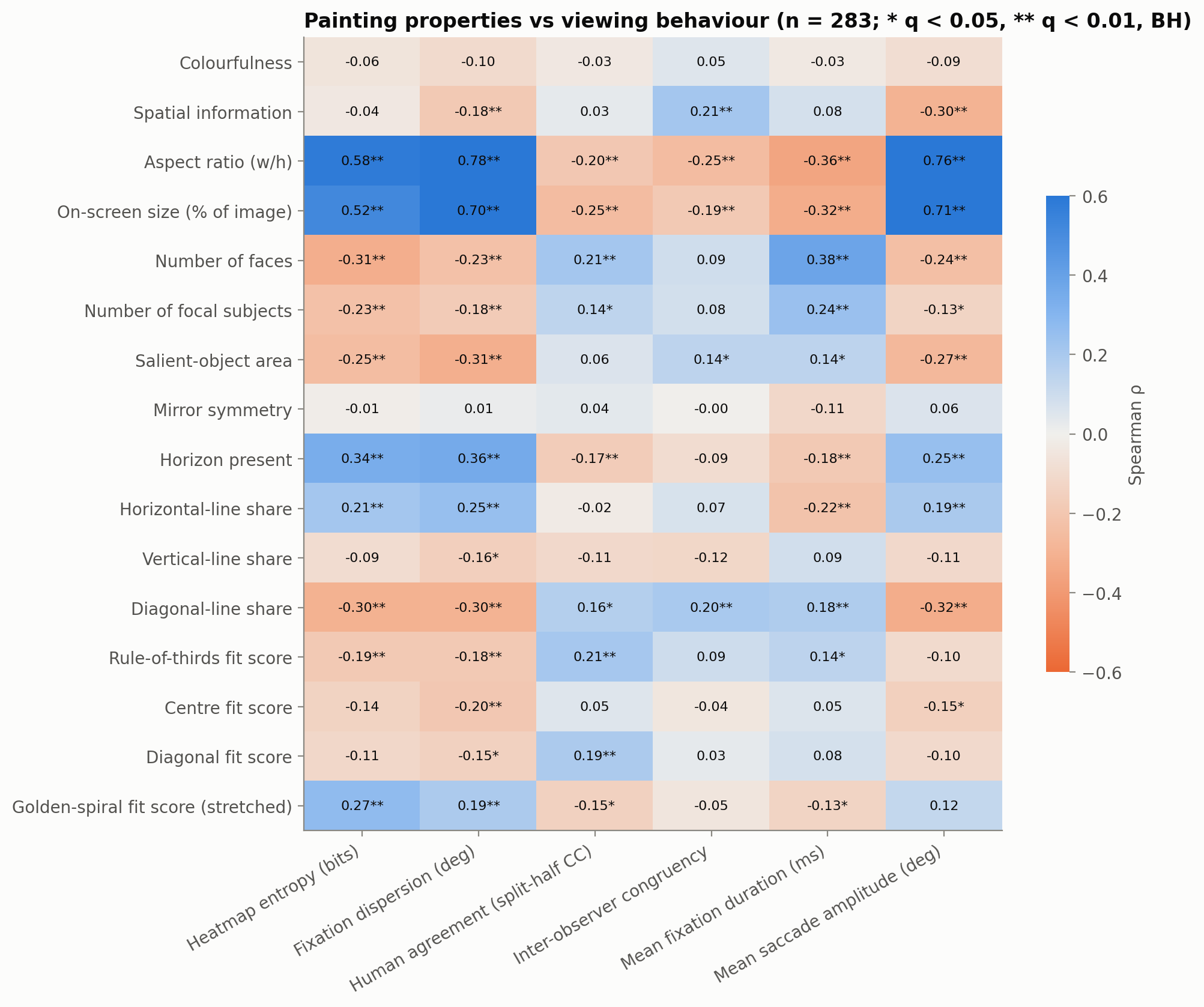}
  \caption{Painting properties vs viewing behaviour (Spearman $\rho$; * $q<0.05$, ** $q<0.01$).}
  \label{fig:supp-factors}
\end{figure}

\begin{figure}[h]
  \centering
  \includegraphics[width=\textwidth]{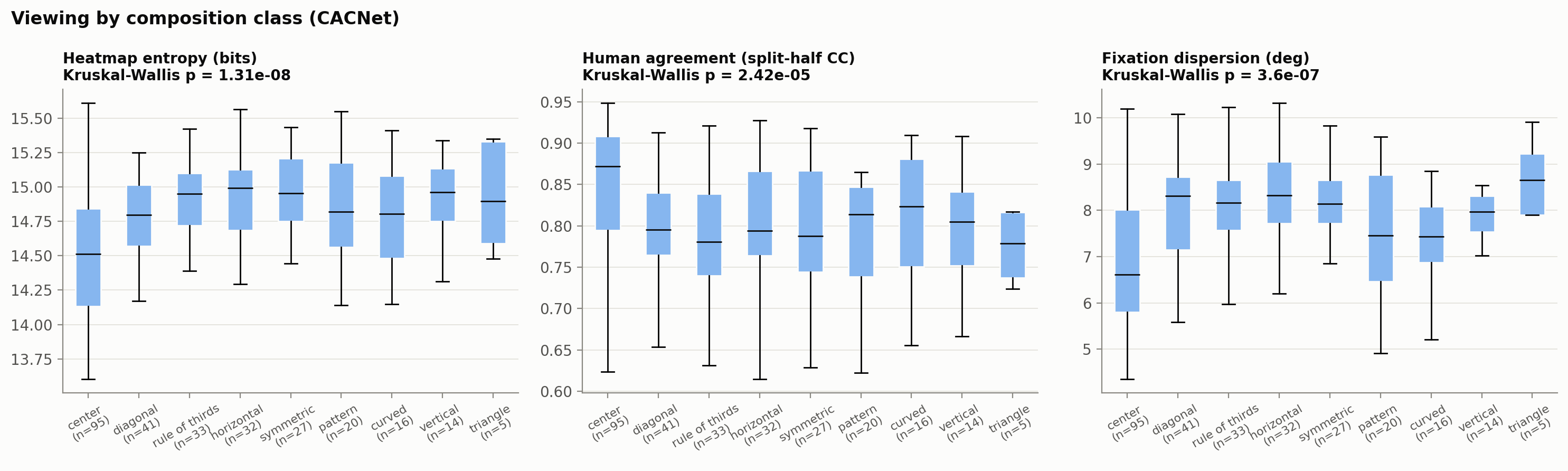}
  \caption{Viewing by CACNet composition class.}
\end{figure}

\section{Style analysis}

\begin{figure}[h]
  \centering
  \includegraphics[width=0.75\textwidth]{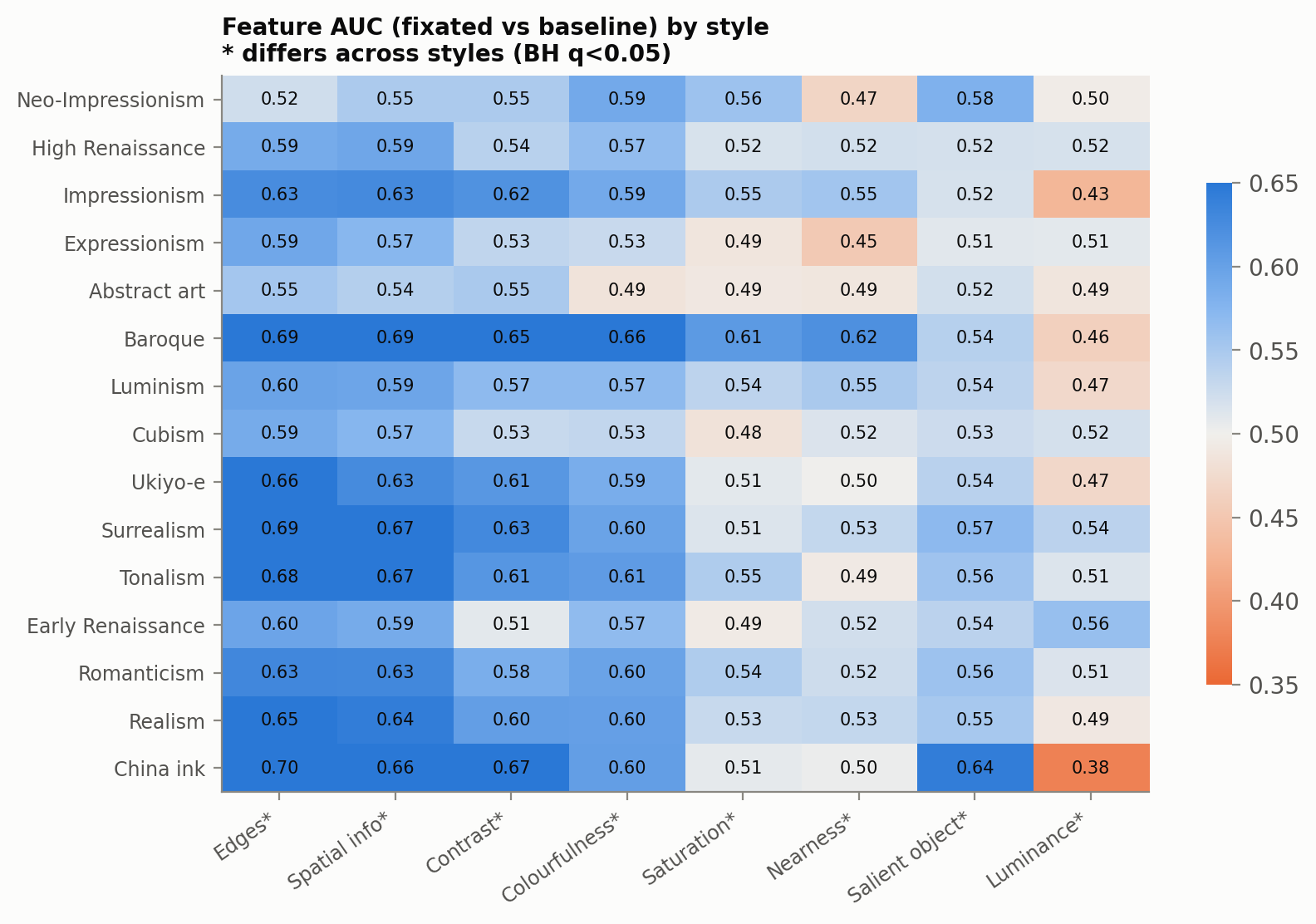}
  \caption{Image-feature AUC (fixated vs baseline) by style.}
  \label{fig:supp-style-features}
\end{figure}

All viewing measures in this section use the first 5\,s and are adjusted for exposure duration. Movement groups with
fewer than five paintings are omitted. Feature AUCs (\cref{fig:supp-style-features}) are computed over the full
viewing, so differences in exposure contribute to them.

\begin{figure}[h]
  \centering
  \includegraphics[width=\textwidth]{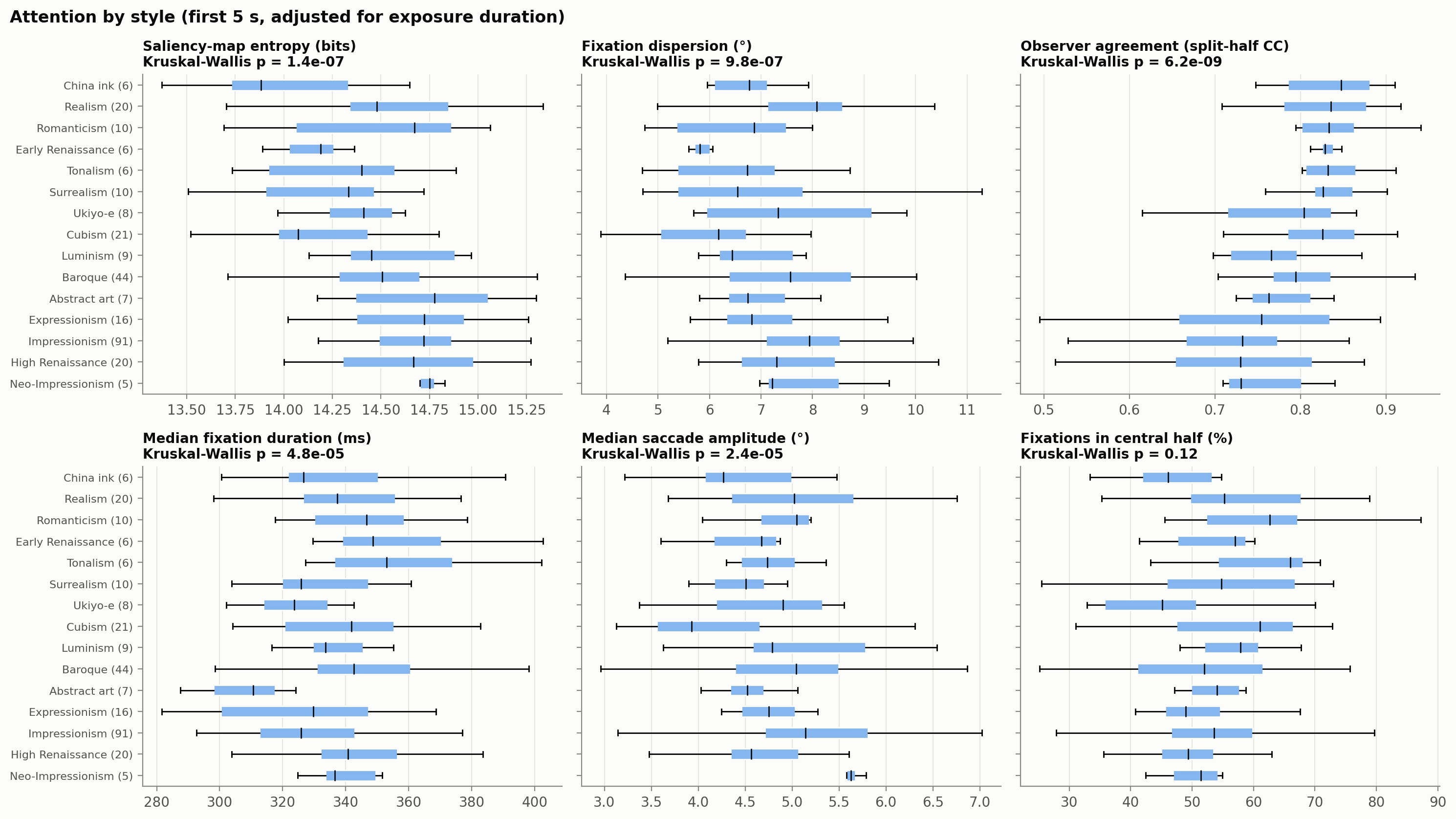}
  \caption{Viewing measures by style (number of paintings in parentheses).}
  \label{fig:supp-style-attention}
\end{figure}

\begin{figure}[h]
  \centering
  \includegraphics[width=\textwidth]{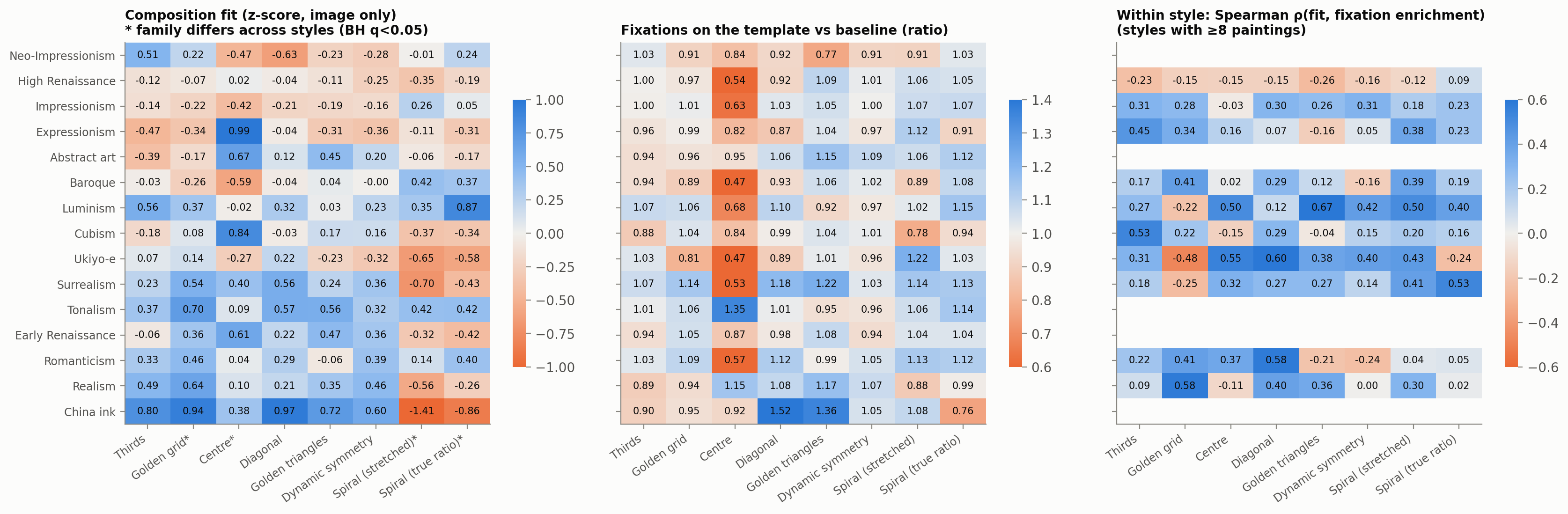}
  \caption{Composition by style. Left: mean image-only fit (z-scored; * family differs across styles). Middle: fixation
  enrichment on the template. Right: within-style correlation between fit and fixation enrichment.}
  \label{fig:supp-style-composition}
\end{figure}

\begin{figure}[h]
  \centering
  \includegraphics[width=\textwidth]{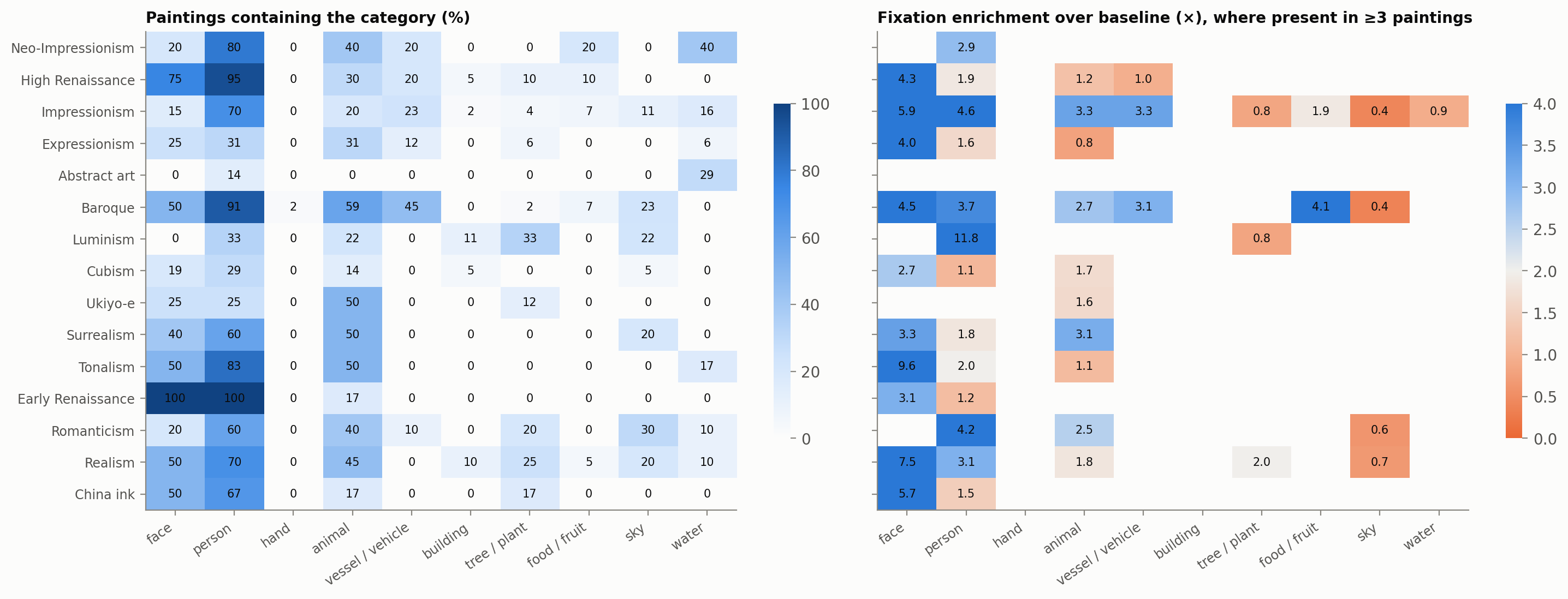}
  \caption{Objects by style: share of paintings containing each category (left) and fixation enrichment over the
  baseline where present (right).}
  \label{fig:supp-style-objects}
\end{figure}

\end{document}